# Kalman Temporal Differences

**Matthieu Geist**                                    MATTHIEU.GEIST@SUPELEC.FR

**Olivier Pietquin**                                  OLIVIER.PIETQUIN@SUPELEC.FR

*IMS research group*

*Supélec*

*Metz, France*

## Abstract

Because reinforcement learning suffers from a lack of scalability, online value (and $Q$-) function approximation has received increasing interest this last decade. This contribution introduces a novel approximation scheme, namely the *Kalman Temporal Differences* (KTD) framework, that exhibits the following features: sample-efficiency, non-linear approximation, non-stationarity handling and uncertainty management. A first KTD-based algorithm is provided for deterministic Markov Decision Processes (MDP) which produces biased estimates in the case of stochastic transitions. Than the eXtended KTD framework (XKTD), solving stochastic MDP, is described. Convergence is analyzed for special cases for both deterministic and stochastic transitions. Related algorithms are experimented on classical benchmarks. They compare favorably to the state of the art while exhibiting the announced features.

## 1. Introduction

Optimal control of stochastic dynamic systems is a trend of research with a long history. The machine learning response to this recurrent problem is the Reinforcement Learning (RL) paradigm (Bertsekas & Tsitsiklis, 1996; Sutton & Barto, 1998; Sigaud & Buffet, 2010). In this general paragon, an artificial agent learns an optimal control policy through interactions with the dynamic system (also considered as its environment). After each interaction, the agent receives an immediate scalar reward information and the optimal policy it searches for is the one that maximizes the cumulative reward over the long run.

Traditionally the dynamic system to be controlled is modeled as a Markov Decision Process (MDP). An MDP is a tuple $\{S, A, P, R, \gamma\}$, where $S$ is the state space, $A$ the action space, $P : s, a \in S \times A \rightarrow p(.|s,a) \in \mathcal{P}(S)$ the family of transition probabilities, $R : S \times A \times S \rightarrow \mathbb{R}$ the bounded reward function, and $\gamma$ the discount factor (weighting long-term rewards). According to these definitions, the system stochastically steps from state to state conditionally on the actions the agent performed. To each transition $(s_i, a_i, s_{i+1})$ is associated an immediate reward $r_i$. A policy $\pi : S \rightarrow A$ is a mapping from states to actions which drives the action selection process of the agent. The optimal policy $\pi^*$ is the one that maximizes the cumulative reward over the long term.

This cumulative reward is locally estimated by the agent as a so-called value (respectively $Q$-) function associating an expected cumulative reward to each state (respectively state-action pair). The optimal policy is therefore the one that maximizes these functions for each state or state-action pair. Many RL algorithms aim at estimating one of these functions so as to infer the optimal policy. In the more challenging cases, the search for the optimal policy is done online, while controlling the system. This requires a trial and error process and a dilemma between immediate exploitation of the currently learnt policy and exploration to improve the policy then occurs.

In this context, a fair RL algorithm should address some important features:

- allowing online learning;

- handling large or even continuous state spaces;

- being sample-efficient (learning a good control policy from as few interactions as possible);

- dealing with non-stationarity (even if the system is stationary, controlling it while learning the optimal policy induces non-stationarities; other good reasons to prefer tracking to convergence are given in Sutton, Koop, & Silver, 2007);

- managing uncertainty (which is a useful information for handling the dilemma between exploration and exploitation);

- handling non-linearities (to deal with the max operator of the Bellman optimality equation and for compact function representations such as neural networks).

All these aspects are rarely addressed at the same time by state-of-the-art RL algorithms. We show that the proposed *Kalman Temporal Differences* (KTD) framework (Geist, Pietquin, & Fricout, 2009a) addresses all these issues. It is based on the Kalman filtering paradigm and uses an approximation scheme, namely the Unscented Transform (UT) of Julier and Uhlmann (2004), to approximate the value function. Originally the Kalman (1960) filtering paradigm aims at tracking the hidden state (modeled as a random variable) of a non-stationary dynamic system through indirect observations of this state. The idea underlying KTD is to cast value function approximation into a filtering problem, so as to benefit from intrinsic advantages of Kalman filtering: online second order learning, uncertainty estimation and non-stationarity handling. The UT is used to deal with non-linearities in a derivative-free fashion, which notably allows deriving a second-order value iteration-like algorithm (namely KTD-Q).

## 1.1 Formalism

The value function $V^\pi$ of a given policy $\pi$ associates to each state the expected discounted cumulative reward for starting in this state and then following $\pi$:

$$V^\pi(s) = E[\sum_{i=0}^{\infty} \gamma^i r_i | s_0 = s, \pi]  \tag{1}$$

where $r_i$ is the reward observed at time $i$. The $Q$-function adds a degree of freedom for the choice of the first action:

$$Q^\pi(s, a) = E[\sum_{i=0}^{\infty} \gamma^i r_i | s_0 = s, a_0 = a, \pi]  \tag{2}$$

Reinforcement learning aims at finding (through interactions) the policy $\pi^*$ which maximises the value function for every state:

$$\pi^* = \operatorname*{argmax}_{\pi}(V^\pi) \qquad (3)$$

Despite the partial order (value functions are vectors), this maximum exists (Puterman, 1994). Two schemes (among others) can lead to the solution. First, *policy iteration* implies learning the value function of a given policy, then improving the policy, the new one being greedy respectively to the learnt value function. It requires solving the *Bellman evaluation equation* (given here for the value function and the $Q$-function):

$$V^\pi(s) = E_{s'|s,\pi(s)}\left[R(s,\pi(s),s') + \gamma V^\pi(s')\right], \forall s \in S \qquad (4)$$

$$Q^\pi(s,a) = E_{s'|s,a}\left[R(s,a,s') + \gamma Q^\pi(s',\pi(s'))\right], \forall s,a \in S \times A \qquad (5)$$

The expectations depend on the transition probability conditioned on current state-action pair, the action being given by the policy in the case of value function evaluation. The second scheme, called *value iteration*, aims at directly finding the optimal policy. It requires solving the *Bellman optimality equation* (given here for the $Q$-function):

$$Q^*(s,a) = E_{s'|s,a}\left[R(s,a,s') + \gamma \max_{b \in A} Q^*(s',b)\right], \forall s,a \in S \times A \qquad (6)$$

A parametric representation of either the value or the $Q$-function is supposed to be available (possible representations are discussed hereafter) and Temporal Differences (TD) algorithms are considered. TD algorithms form a class of online methods which consist in correcting the representation of the value (or $Q$-) function according to the so-called TD error $\delta_i$ made on it. Although the formal definition of the TD error depends on the algorithm (see Section 1.2), it can be intuitively defined as the difference between the predicted reward according to the current estimate of the value or $Q$-function and the actual observed reward at time step $i$. Most of TD algorithms can be generically written as:

$$\theta_i = \theta_{i-1} + K_i \delta_i \qquad (7)$$

In this expression, $\theta_{i-1}$ is the latest estimate of the value function (or of the set of parameters defining it), $\theta_i$ is an updated representation given an observed transition, $\delta_i$ is the TD error, and $K_i$ is a gain indicating the direction in which the representation of the target function should be corrected.

If the state space $S$ and the action space $A$ are finite and small enough, an exact description of the value function is possible, and $\theta$ is a vector with as many components as the state (-action) space (tabular representation). In the case of large state and/or action spaces, approximation is necessary. A classical choice in RL is the linear parameterization, that is the value function is approximated by:

$$\hat{V}_\theta(s) = \sum_{j=1}^{p} w_j \phi_j(s) = \phi(s)^T \theta \qquad (8)$$

where $(\phi_j)_{1 \leq j \leq p}$ is a set of basis functions, which should be defined beforehand, and the weights $w_j$ are the parameters:

$$\theta = \begin{pmatrix} w_1 & \dots & w_p \end{pmatrix}^T \text{ and } \phi(s) = \begin{pmatrix} \phi_1(s) & \dots & \phi_p(s) \end{pmatrix}^T \tag{9}$$

Many function approximation algorithms require such a representation to ensure convergence (Tsitsiklis & Roy, 1997; Schoknecht, 2002), or even to be applicable (Bradtke & Barto, 1996; Boyan, 1999; Geramifard, Bowling, & Sutton, 2006). Other representations are possible such as neural networks where $\theta$ is the set of synaptic weights (usually resulting in a nonlinear dependency of the value function to its parameters).

Adopting this generic point of view, the problem addressed in this paper can be stated as: given a representation of the value function (or of the $Q$-function) summarized by the parameter vector $\theta$ and given a Bellman equation to be solved, what is the "best" gain $K$? Some state-of-the-art answers to this question are given in the following section.

## 1.2 State of the Art

This paper focuses on online methods. Standard RL algorithms such as TD evaluation, SARSA and Q-Learning (Sutton & Barto, 1998) share the same features and a unified view based on Equation (7) is adopted in the following. In this equation, the term $\delta_i$ is the TD error. Suppose that at step $i$ a transition $(s_i, a_i, r_i, s_{i+1}, a_{i+1})$ is observed. For TD-like RL algorithms, that is algorithms aiming at evaluating the value function of a given policy $\pi$, the TD error is:

$$\delta_i = r_i + \gamma \hat{V}_{\theta_{i-1}}(s_{i+1}) - \hat{V}_{\theta_{i-1}}(s_i) \tag{10}$$

For SARSA-like algorithms, that is algorithms which aim at evaluating the $Q$-function of a given policy $\pi$, the TD error is:

$$\delta_i = r_i + \gamma \hat{Q}_{\theta_{i-1}}(s_{i+1}, a_{i+1}) - \hat{Q}_{\theta_{i-1}}(s_i, a_i) \tag{11}$$

Finally, for $Q$-learning-like algorithms, that is algorithms which aim at computing the optimal $Q$-function $Q^*$, the TD error is:

$$\delta_i = r_i + \gamma \max_{b \in A} \hat{Q}_{\theta_{i-1}}(s_{i+1}, b) - \hat{Q}_{\theta_{i-1}}(s_i, a_i) \tag{12}$$

The type of temporal difference determines the Bellman equation to be solved (evaluation equation for (10-11), optimality equation for (12)), and thus if the algorithm belongs to the policy iteration or value iteration family.

The gain $K_i$ is specific to each algorithm. The most common are reviewed here. For TD, SARSA and $Q$-learning (for example, see Sutton & Barto, 1998), the gain can be written as

$$K_i = \alpha_i e_i \tag{13}$$

where $\alpha_i$ is a classical learning rate in stochastic approximation theory which should satisfy:

$$\sum_{i=0}^{\infty} \alpha_i = \infty \text{ and } \sum_{i=0}^{\infty} \alpha_i^2 < \infty \tag{14}$$

and $e_i$ is a unitary vector which is zero everywhere except in the component corresponding to state $s_i$ (or to state-action $(s_i, a_i)$) where it is equal to one (Kronecker function). These algorithms have been modified to consider so-called eligibility traces (again, see Sutton and Barto), and the gain is then written as

$$K_i = \alpha_i \sum_{j=1}^{i} \lambda^{i-j} e_j \tag{15}$$

where $\lambda$ is the eligibility factor. Informally, this approach keeps memory of trajectories in order to propagate updates to previously visited states.

These algorithms have also been extended to take into account approximate representation of the value function (Sutton & Barto, 1998), and are called direct algorithms (Baird, 1995). Without eligibility traces, the gain is written as

$$K_i = \alpha_i \nabla_{\theta_{i-1}} \hat{V}_{\theta_{i-1}}(s_i) \tag{16}$$

where $\nabla_{\theta_{i-1}} \hat{V}_{\theta_{i-1}}(s_i)$ is the gradient following the parameter vector of the parameterized value function in the current state. This gain corresponds to a stochastic gradient descent according to the cost function $\|V^\pi - \hat{V}_\theta\|^2$. As $V^\pi(s_i)$ is not known nor directly observable, it is replaced by $r_i + \gamma \hat{V}_\theta(s_{i+1})$. This general approach is known as bootstrapping (Sutton & Barto, 1998). The value function can be replaced straightforwardly by the $Q$-function in this gain. The direct algorithms have also been extended to take into account eligibility traces, which leads to the following gain:

$$K_i = \alpha_i \sum_{j=1}^{i} \lambda^{i-j} \nabla_{\theta_{i-1}} \hat{V}_{\theta_{i-1}}(s_j) \tag{17}$$

Another well known approach is the set of residual algorithms (Baird, 1995), for which the gain is obtained through the minimization of the $L_2$-norm of the Bellman residual (*i.e.*, the difference between the left side and the right side of the Bellman equation, possibly for sampled transitions) using a stochastic gradient descent:

$$K_i = \alpha_i \nabla_{\theta_{i-1}} \left( \hat{V}_{\theta_{i-1}}(s_i) - \gamma \hat{V}_{\theta_{i-1}}(s_{i+1}) \right) \tag{18}$$

The next reviewed approach is the (recursive form of the) Least-Squares Temporal Differences (LSTD) algorithm of Bradtke and Barto (1996), which is only defined for a linear parameterization (8) and for which the gain is defined recursively:

$$K_i = \frac{C_{i-1} \phi(s_i)}{1 + (\phi(s_i) - \gamma \phi(s_{i+1}))^T C_{i-1} \phi(s_i)} \tag{19}$$

$$C_i = C_{i-1} - \frac{C_{i-1} \phi(s_i)(\phi(s_i) - \gamma \phi(s_{i+1}))^T C_{i-1}}{1 + (\phi(s_i) - \gamma \phi(s_{i+1}))^T C_{i-1} \phi(s_i)} \tag{20}$$

where $\phi(s)$ is defined in (9) and for which the matrix $C_0$ must be initialized. LSTD also seeks to minimize the $L_2$-norm of the Bellman residual, however using a least-squares approach rather than a gradient descent and using the instrumental variable concept (Söderström

& Stoica, 2002) to cope with stochasticity of transitions[1]. This algorithm has also been extended to eligibility traces (for details, see Boyan, 1999).

The last reviewed approach, which is certainly the closest to this contribution, is the Gaussian Process Temporal Differences (GPTD) algorithm of Engel (2005). A linear parameterization $V_\theta(s) = \phi(s)^T\theta$ is assumed[2] and the following statistical generative model (obtained from the Bellman evaluation equation) is considered:

$$\begin{pmatrix} r_1 \\ \vdots \\ r_i \end{pmatrix} = \begin{pmatrix} 1 & -\gamma & 0 & \cdots \\ 0 & 1 & -\gamma & 0 \\ \vdots & \ddots & \ddots & -\gamma \\ 0 & \cdots & 0 & 1 \end{pmatrix} \begin{bmatrix} \phi(s_1)^T \\ \vdots \\ \phi(s_i)^T \end{bmatrix} \theta + \begin{pmatrix} n_1 \\ \vdots \\ n_i \end{pmatrix} \tag{21}$$

By assuming that the noise $n_j$ is white (and therefore centered), Gaussian and of variance $\sigma_j$, and that the prior over parameters follows a normal distribution, the posterior distribution of $(\theta|r_1, \ldots, r_i)$ can be analytically computed. Moreover, by using the Sherman-Morrison formula, a recursive algorithm satisfying the Widrow-Hoff update rule (7) can be obtained (assuming a prior $P_0$):

$$K_i = \frac{P_{i-1}(\phi(s_i) - \gamma\phi(s_{i+1}))}{\sigma_i^2 + (\phi(s_i) - \gamma\phi(s_{i+1}))^T P_{i-1}(\phi(s_i) - \gamma\phi(s_{i+1}))} \tag{22}$$

$$P_i = P_{i-1} - \frac{P_{i-1}(\phi(s_i) - \gamma\phi(s_{i+1}))(\phi(s_i) - \gamma\phi(s_{i+1}))^T P_{i-1}}{\sigma_i^2 + (\phi(s_i) - \gamma\phi(s_{i+1}))^T P_{i-1}(\phi(s_i) - \gamma\phi(s_{i+1}))} \tag{23}$$

Alternatively, GPTD (with parametric representation) can be seen as the linear least-squares solution of the $L_2$ Bellman residual minimization.

Only the most classical value function approximation algorithms have been presented, however many other exist. Nevertheless, to our knowledge none of them presents all the features argued before as being desirable. Most of them assumes linearity, at least to ensure convergence (Tsitsiklis & Roy, 1997; Schoknecht, 2002) and sometime even to be applicable (Bradtke & Barto, 1996; Boyan, 1999; Geramifard et al., 2006). Some other algorithms do not assume linearity, as residual ones (Baird, 1995), however they are not often practical (eg., a value iteration-like residual algorithm is proposed by Baird, but this method requires computing the gradient of the max operator). Some of these methods are more sample efficient than others. Generally speaking, second order approaches tend to be more efficient than first order one, and LSTD is usually recognized as being a sample efficient approach. Algorithms which use a learning rate can partially cope with non-stationarity, by using an adaptive learning rate for example. However the LSTD approach is known to not take into account non-stationary (which explains that it is almost never used in optimistic policy iteration or incremental actor-critic schemes), see for example the work of Phua and Fitch (2007). Many recent approaches for handling the dilemma between exploration and exploitation use some uncertainty information (eg., see Dearden, Friedman, & Russell, 1998 or Strehl, Li, Wiewiora, Langford, & Littman, 2006). However, as far as we know, very few algorithms allow providing uncertainty information within a value approximation context, and among them is the GPTD framework of Engel (2005). However, contrary to this contribution the effective use of this information is left for future work. Like LSTD, GPTD algorithms are sample efficient but they do not handle non-stationary[3]. Yet, GPTD and KTD frameworks share some similarities, this is discussed throughout this paper. The motivation behind KTD is to handle all these aspects at the same time.

---

1. This point of view is historical. Since then, it has been shown that LSTD actually minimizes the distance between the value function and the projection onto the hypothesis space of its image through the Bellman operator (Lagoudakis & Parr, 2003).

2. Actually, Engel's work is more general. It models the value function itself as a Gaussian process and uses a dictionary method to obtain a sparse representation (without this procedure, the value function would be represented as a vector with as many components as visited states). However, if this dictionary method is used in a preprocessing step, the Gaussian process nonparametric representation reduces to the proposed parametric linear representation, basis functions being kernels. Constructing the parameterization automatically and online is surely of interest, but the proposed point of view makes further comparisons easier.

### 1.3 Paper Outline

The next section introduces an alternative point of view of value function approximation and introduces informally Kalman filtering and the state-space representation, upon which our contribution is built.

Determinism of MDP is assumed in Section 3 and the general Kalman Temporal Differences framework is derived. Deterministic transitions are to be linked to a white noise assumption which is necessary to KTD derivation. It is then specialized using an approximation scheme, the Unscented Transform (UT) of Julier and Uhlmann (2004) to derive a family of practical algorithms. In Section 4, a colored noise model initially introduced by Engel, Mannor, and Meir (2005) is used to extend the KTD framework to the case of stochastic transitions. An eXtended KTD (XKTD) framework is proposed, and its combination with off-policy learning is discussed. Convergence is analysed in Section 5. Under white noise assumption, it is shown that KTD minimizes a weighted square Bellman residual. Under colored noise assumption, it is shown that XKTD indeed performs a least-squares supervised learning associating state values to observed Monte Carlo returns of cumulative rewards. This is the same solution as LSTD(1), which is an unbiased estimator of the value function. Section 6 shows how to compute uncertainty about value estimates from this framework and introduces a form of active learning scheme which aims at improving speed of convergence of KTD-Q, the KTD value iteration-like algorithm. The proposed framework is then experimented and compared to state of the art RL algorithms. Each experiment is a classic RL benchmark which aims at highlighting a specific features of KTD. Last section discusses position of the proposed framework to other related approaches and offers some perspectives.

## 2. An Alternative Point of View

The previous section presented the standard vision of the reinforcement learning problem and of its formulation under the MDP framework. Here an alternative point of view is introduced.

---

3. LSTD and GPTD could certainly be extended to the non-stationary case, for example by introducing some forgetting factor. However, this is not how they have been designed initially, and the aim of this paper is not to provide LSTD nor GPTD variations.

## 2.1 Informal Idea

In this paper, a novel approach based on an alternative point of view is proposed. A stochastic dynamic system is seen as possessing underlying value functions $V \in \mathbb{R}^S$ and state-action value functions $Q \in \mathbb{R}^{S \times A}$ that an agent can observe by interacting with the system. When an agent takes an action, it provokes a state change and the generation of a reward. This reward is actually a local observation of the set of underlying value functions ruling the behavior of the system. From a sequence of such observations, the agent can infer information about any of the value functions. A good estimate of the value function $\hat{V}(s)$ (resp. state-action value function $\hat{Q}(s,a)$) is given by the conditional expectation over all possible trajectories of $V(s)$ (resp. $Q(s,a)$) given the sequence of observed rewards:

$$\hat{V}_i(s) = E[V(s)|r_1, \ldots, r_i] \tag{24}$$

$$\hat{Q}_i(s,a) = E[Q(s,a)|r_1, \ldots, r_i] \tag{25}$$

Interacting with the system therefore becomes a mean to generate observations that helps estimating value functions which are hidden properties of the system. From these value function estimates, the followed policy can be modified to move towards the optimal policy. It is also legitimate to adopt a behavior that allows gathering meaningful observations which relates to the exploration versus exploitation dilemma.

Two special cases of value functions are the one associated to the followed policy $\pi$ and the one associated to the optimal policy $\pi^*$. The rest of this paper concentrates on estimating these two particular value functions or associated $Q$-functions.

Equations (24) and (25) are not solvable in the general case but inferring hidden variables from observations is typically treated by Kalman filtering in the signal processing and optimal control communities. Value functions will be considered as generated by a set of parameters and the search is for the optimal set of hidden parameters $\theta^*$ that provides the best estimate of the value function (see Section 3.1). In the following, Kalman filtering is first introduced and a method casting (state-action) value function approximation into the Kalman filtering framework and using Bellman equations to build a so-called state-space representation of the problem is proposed.

## 2.2 Kalman Filtering

Originally, the Kalman (1960) filtering paradigm aims at tracking the hidden state $X$ (modeled as a random vector) of a non-stationary dynamic system through indirect observations $\{Y_1, \ldots, Y_i\}$ of this state. To do so, at time $i-1$ the algorithm computes a prediction of the state ($\hat{X}_{i|i-1}$) and observation ($\hat{Y}_{i|i-1}$) at time $i$, knowing analytically how states evolve and generate observations as clarified below. After the actual next observation $Y_i$ is known (at time $i$), the state prediction is corrected to obtain the state estimate $\hat{X}_{i|i}$ using the observation prediction error ($e_i = Y_i - \hat{Y}_{i|i-1}$) according to the following Windrow-Hoff-like equation:

$$\hat{X}_{i|i} = \hat{X}_{i|i-1} + K_i(Y_i - \hat{Y}_{i|i-1}) = \hat{X}_{i|i-1} + K_i e_i \tag{26}$$

where $K_i$ is the *Kalman gain* which will be further described hereafter. In the original work of Kalman, the linear form of equation (26) is a constraint: adopting a statistical point of view, the goal of the Kalman filter is to recursively compute the *best linear estimate* $\hat{X}_i$ of the state at time $i$ given the sequence of observations $\{Y_1, \ldots, Y_i\}$. Kalman considers the best estimate to be the one that minimizes the quadratic cost function

$$J_i(\hat{X}) = E[\|X_i - \hat{X}\|^2 | Y_1, \ldots, Y_i] \tag{27}$$

To compute the optimal gain $K_i$ under the constraints (26) and (27), several assumptions are made.

First, the evolution of the system is supposed to be ruled by a so-called *evolution equation* or *process equation* (using the possibly non-stationary $f_i$ function) which is known:

$$X_{i+1} = f_i(X_i) + v_i \tag{28}$$

Equation (28) links the next state $X_{i+1}$ with the current one $X_i$ and $v_i$ is a random noise usually named *evolution noise* or *process noise* modeling the uncertainty in the evolution.

Second, observations are supposed to be linked to states by another known function $g_i$ used in the typically called *observation equation* or *sensing equation*:

$$Y_i = g_i(X_i) + w_i \tag{29}$$

Equation (29) relates the current observation $Y_i$ to the current state $X_i$ and $w_i$ is a random noise usually named *observation noise* modeling the uncertainty induced by the noisy observation. This noise together with the process noise are at the origin of the state estimation problem (estimating the current state from history of observations).

Equations (28) and (29) provide the so-called *state-space* description of the system. The major assumptions of Kalman is that $v_i$ and $w_i$ are additive, white and independent noises of variance $P_v$ and $P_w$ respectively, meaning that:

$$E[v_i] = E[w_i] = 0 \tag{30}$$
$$E[v_i \cdot w_j] = 0 \quad \forall i, j \tag{31}$$
$$E[v_j \cdot v_i] = E[w_j \cdot w_i] = 0 \quad \forall i \neq j \tag{32}$$

Given these assumptions and the constrains (26) and (27) and adopting a statistical point of view, the Kalman filter algorithm provides the optimal quantities $\hat{X}_{i|i-1}$, $\hat{Y}_{i|i-1}$ and $K_i$:

$$\hat{X}_{i|i-1} = E[X_i | Y_1, \ldots, Y_{i-1}] = E[f_{i-1}(X_{i-1}) + v_{i-1} | Y_1, \ldots, Y_{i-1}]$$
$$= E[f_{i-1}(X_{i-1}) | Y_1, \ldots, Y_{i-1}] = E[f_{i-1}(\hat{X}_{i-1|i-1})], \tag{33}$$
$$\hat{Y}_{i|i-1} = E[Y_i | Y_1, \ldots, Y_{i-1}] = E[g_i(X_i) + w_i | Y_1, \ldots, Y_{i-1}]$$
$$= E[g_i(X_i) | Y_1, \ldots, Y_{i-1}] = E[g_i(\hat{X}_{i-1|i-1})], \tag{34}$$
$$K_i = P_{Xe_i} P_{e_i}^{-1}. \tag{35}$$

where $P_{Xe_i} = E[(X_i - \hat{X}_{i|i-1})e_i | Y_1, \ldots, Y_{i-1}]$ and $P_{e_i} = \text{cov}(e_i | Y_1, \ldots, Y_{i-1})$.

It is not in the scope of this paper to provide the complete development leading to these general results which are provided by Kalman (1960). Yet, Section 3 will provide further developments in the specific case of RL.

Several important comments can be made at this stage. First, no specific assumption has been made about the distributions of the noises $v$ and $w$ except that they have a zero-mean and known variances ($P_v$ and $P_w$). Given this, the Kalman filter provides the best linear estimator (in the sense that the estimator's update rule is linear) of the system's state which may not be optimal. Yet, if these two noises have Gaussian distributions, they are totally described by their mean and variance. In this specific case, the linear estimate is thus the optimal estimate and the Kalman filter algorithm provides the optimal solution. In this paper, the Gaussian assumption is never made and only the best linear estimator is considered.

Second, no linear assumption has been made concerning functions $f_i$ and $g_i$. Although Kalman (1960) provides exact solutions to the estimation problem in the case of linear state-space equations, only quantities involved in (33), (34) and (35) are required. There exists approximation schemes to estimate these quantities even in the case of non-linear equations. Extended Kalman filters and the unscented transform (see Section 3.2.2) are such schemes.

Finally, Kalman filtering should not be mistaken for Bayesian filtering. Bayesian filtering would consist in computing the complete posterior distribution of the state given the observations. Kalman filtering only focuses on the first and second moments of this distribution (mean and variance) with a constrained linear update. In the case of Gaussian distributions, Bayesian filtering reduces to Kalman filtering but is more complex in the general case. In this paper, only Kalman filtering is considered.

## 2.3 State-space Formulation for the Value Function Evaluation Problem

Before providing the general framework, underlying ideas are introduced through the value function $V^\pi(s)$ evaluation problem. As providing some uncertainty information about estimates is considered as a desired feature, a statistical point of view is adopted and the parameter vector $\theta$ is modeled as a set of random variables. Another desired feature is to track the solution rather than converging to it. This suggests adopting some evolution model for the value function (through the parameters). However, dynamics of the value function are hard to model, as they depend on whether the dynamic system to be controlled is non-stationary or the value function evaluation takes place in a generalized policy iteration scheme[4]. Here a heuristic evolution model following the Occam razor principle is adopted and parameters evolution is modeled as a random walk:

$$\theta_i = \theta_{i-1} + v_i \tag{36}$$

In this equation, $\theta_i$ is the (true) parameter vector at time $i$ and $v_i$ is the evolution noise. It is assumed white (that is centered, and at two different time steps, noises are independent), but no hypothesis is done about its distribution. The parameter vector $\theta_i$ is thus a random process. As it is stationary (because $E[\theta_i] = E[\theta_{i-1}]$), it should not harm the case where the value function is stationary. On the other hand, it should allow tracking a non-stationary value function (even if this evolution model is not the true one, which cannot anyway be obtained in the general case).

---

4. Each time the policy is improved, the associated value function changes too. Therefore, the value function to be learnt is non-stationary.

Another issue is to link what is observed (the reward) to what needs to be inferred (the parameter vector representing the value function). The Bellman evaluation equation is a good candidate to produce such an observation model:

$$r_i = V^\pi(s_i) - \gamma V^\pi(s_{i+1}) \tag{37}$$

However, the solution of the Bellman equation does not necessarily lie in the hypothesis space (the set of functions which can be represented by the parameter vector, for a given representation). Therefore there is some inductive bias $n_i$, which is modeled here as a centered noise:

$$r_i = \hat{V}_{\theta_i}(s_i) - \gamma \hat{V}_{\theta_i}(s_{i+1}) + n_i \tag{38}$$

Notice again that no Gaussian assumption is made about the distribution of this noise.

Evolution and observation models can be summarized in the following "state-space formulation":

$$\begin{cases} \theta_i & = \theta_{i-1} + v_i \\ r_i & = \hat{V}_{\theta_i}(s_i) - \gamma \hat{V}_{\theta_i}(s_{i+1}) + n_i \end{cases} \tag{39}$$

This is a model of value function approximation. It is assumed that there exists some parameter random process $\theta_i$ which generates the rewards through the Bellman evaluation equation, these observations being noisy due to some inductive bias and to the fact that a "sampled" Bellman equation is used instead of the true one. States and actions can be considered here as exogenous variables which are part of the definition of the observation model at time $i$. Estimating the value function reduces here to the estimation of this hidden random process. It can be addressed by Bayesian filtering, which aims at estimating the whole distribution of $\theta_i$ conditioned on past observed rewards. In this paper a more restrictive point of view is adopted, the Kalman filtering one, and only mean and variance of this distribution are estimated with a restriction to linear update rules.

## 3. KTD: the Deterministic Case

From now on and through the rest of this section the focus is on deterministic Markov decision processes. Transitions become deterministic and Bellman equations (4-6) simplify as follows:

$$V^\pi(s) = R(s, \pi(s), s') + \gamma V^\pi(s'), \, \forall s \tag{40}$$

$$Q^\pi(s, a) = R(s, a, s') + \gamma Q^\pi(s', \pi(s')), \, \forall s, a \tag{41}$$

$$Q^*(s, a) = R(s, a, s') + \gamma \max_{b \in A} Q^*(s', b), \, \forall s, a \tag{42}$$

In this section are provided the derivation of the most general KTD algorithm as well as specializations to practical implementations.

### 3.1 The General Framework

A very general point of view is adopted now. A transition is generically noted as:

$$t_i = \begin{cases} (s_i, s_{i+1}) \\ (s_i, a_i, s_{i+1}, a_{i+1}) \\ (s_i, a_i, s_{i+1}) \end{cases} \tag{43}$$

given that the aim is the value function evaluation, the $Q$-function evaluation or the $Q$-function optimization (in other words, the direct evaluation of the optimal $Q$-function). Similarly, for the same cases, the following shortcuts hold:

$$g_{t_i}(\theta_i) = \begin{cases} \hat{V}_{\theta_i}(s_i) - \gamma \hat{V}_{\theta_i}(s_{i+1}) \\ \hat{Q}_{\theta_i}(s_i, a_i) - \gamma \hat{Q}_{\theta_i}(s_{i+1}, a_{i+1}) \\ \hat{Q}_{\theta_i}(s_i, a_i) - \gamma \max_{b \in A} \hat{Q}_{\theta_i}(s_{i+1}, b) \end{cases} \tag{44}$$

Then all TD errors can be written generically as

$$\delta_i = r_i - g_{t_i}(\theta_i) \tag{45}$$

A statistical point of view is adopted. As said before, the original Kalman (1960) filter paradigm aims at tracking the hidden state (modeled as a random variable) of a non-stationary dynamic system through indirect observations of this state. The idea behind KTD is to express value function approximation as a filtering problem: the parameters are the hidden state to be tracked (modeled as random variables following a random walk), the observation being the reward linked to the parameters through a Bellman equation. The problem at sight can then be stated in a so-called *state-space formulation* (this term comes from Kalman filtering literature and should not be confused with the state space of an MDP):

$$\begin{cases} \theta_i = \theta_{i-1} + v_i \\ r_i = g_{t_i}(\theta_i) + n_i \end{cases} \tag{46}$$

This expression is fundamental for the proposed framework. Using the vocabulary of Kalman filtering, the first equation is the evolution equation, it specifies that the real parameter vector follows a random walk which expectation corresponds to the optimal estimate of the value function. The evolution noise $v_i$ is white, independent and of variance matrix $P_{v_i}$ (to be chosen by the practitioner, this is further discussed in section 7). Notice that this equation is *not* an update of the parameters (addressed later), but model their natural evolution over time, according to the Kalman filtering paradigm described in Section 2.2; notably this allows handling non-stationarity of the targeted value function. The second equation is the observation equation, it links the observed transition to the value (or $Q$-) function through a Bellman equation, see (44). The observation noise $n_i$ is supposed white, independent and of (scalar) variance $P_{n_i}$ (also to be chosen by the practitioner and further discussed in section 7). Notice that this mandatory assumption does not hold for stochastic MDP, that is why deterministic transitions are supposed here. More details about this assumption and its consequences are given in Section 4. Given deterministic transitions, this model noise arises because the solution of the Bellman equation does not necessarily exists in the hypothesis space induced by the parameterization. Notice that the choice of the nature of the approximator (choice of the structure of a neural network, of basis functions for linear parameterization, *etc.*) is an important topic in reinforcement learning and more generally in machine learning. Nevertheless, it is not addressed here, and it has to be chosen by the practitioner.

### 3.1.1 Minimized Cost Function

An objective could be to estimate the whole distribution of parameters conditioned on past observed rewards, which can be addressed by Bayesian filtering. However, it is a difficult problem in the general case. Here a more simple objective is chosen: estimating the (deterministic) parameter vector which minimizes the expectation over "true" parameters of the mean-squared error conditioned on past observed rewards. The idea is that information is provided by observed transitions and associated rewards, and that knowing the mean of the posterior distribution should be enough. The associated cost can be written as:

$$J_i(\theta) = E\left[\|\theta_i - \theta\|^2 | r_{1:i}\right] \text{ with } r_{1:i} = r_1, \ldots, r_i \tag{47}$$

Notice that if $\theta_i$ is a random vector (of which distribution is not known), $\theta$ is a deterministic vector. Generally speaking, the optimal solution or minimum mean square error (MMSE) estimator is the conditional expectation[5]:

$$\underset{\theta}{\operatorname{argmin}} J_i(\theta) = \hat{\theta}_{i|i} = E\left[\theta_i | r_{1:i}\right] \tag{48}$$

However, except in specific cases, this estimator is not analytically computable. Instead, the aim is here to find the best *linear* estimator of $\theta_i$. It can be written in a form quite similar to equation (7):

$$\hat{\theta}_{i|i} = \hat{\theta}_{i|i-1} + K_i \tilde{r}_i \tag{49}$$

In Equation (49), $\hat{\theta}_{i|i}$ is the estimate of $\theta_i$ at time $i$ and $\hat{\theta}_{i|i-1} = E[\theta_i | r_{1:i-1}]$ is its prediction according to past observed rewards $r_{1:i-1}$, given the evolution equation. For a random walk model the following holds (recall that the evolution noise is white):

$$\hat{\theta}_{i|i-1} = E\left[\theta_{i-1} + v_i | r_{1:i-1}\right] = E\left[\theta_{i-1} | r_{1:i-1}\right]$$
$$= \hat{\theta}_{i-1|i-1} \tag{50}$$

The innovation

$$\tilde{r}_i = r_i - \hat{r}_{i|i-1} \tag{51}$$

is the difference between the actual observed reward $r_i$ and its prediction $\hat{r}_{i|i-1}$ based on the previous estimate of the parameter vector and the observation equation (recall that the observation noise is also white):

$$\hat{r}_{i|i-1} = E\left[r_i | r_{1:i-1}\right] = E\left[g_{t_i}(\theta_i) + n_i | r_{1:i-1}\right]$$
$$= E\left[g_{t_i}(\theta_i) | r_{1:i-1}\right] \tag{52}$$

Note that the innovation $\tilde{r}_i$ is not exactly the temporal difference defined in Equation (45), which is a random variable through its dependency to the random vector $\theta_i$. It is its expectation conditioned on past observed data: $\tilde{r}_i = E[\delta_i | r_{1:i}]$.

---

5. This is quite intuitive, the best deterministic estimator (in a least-squares sens) of a random variable is its mean.

### 3.1.2 Optimal Gain

Using classical equalities, the cost function can be rewritten as the trace of the matrix variance of parameters error:

$$
\begin{aligned}
J_i(\theta) &= E\left[\|\theta_i - \theta\|^2 | r_{1:i}\right] \\
&= E\left[(\theta_i - \theta)^T(\theta_i - \theta) | r_{1:i}\right] \\
&= \text{trace}\left(E\left[(\theta_i - \theta)(\theta_i - \theta)^T | r_{1:i}\right]\right)
\end{aligned} \tag{53}
$$

Recall that we restrict ourselves to the class of linear (and unbiased) estimators depicted in Eq. (49). Therefore, the cost function $J_i(\hat{\theta}_{i|i})$ should be considered, and the unknown is the gain $K_i$:

$$
J_i(\hat{\theta}_{i|i}) = \text{trace}\left(\text{cov}\left(\theta_i - \hat{\theta}_{i|i} | r_{1:i}\right)\right) \tag{54}
$$

A first step to the computation of the optimal gain is to express the conditioned covariance over parameters as a function of the gain $K_i$. A few more notations are first introduced (recall also (51), the definition of the innovation):

$$
\begin{cases}
\tilde{\theta}_{i|i} = \theta_i - \hat{\theta}_{i|i} & \text{and} \quad \tilde{\theta}_{i|i-1} = \theta_i - \hat{\theta}_{i|i-1} \\
P_{i|i} = \text{cov}\left(\tilde{\theta}_{i|i} | r_{1:i}\right) & \text{and} \quad P_{i|i-1} = \text{cov}\left(\tilde{\theta}_{i|i-1} | r_{1:i-1}\right) \\
P_{r_i} = \text{cov}\left(\tilde{r}_i | r_{1:i-1}\right) & \text{and} \quad P_{\theta r_i} = E\left[\tilde{\theta}_{i|i-1} \tilde{r}_i | r_{1:i-1}\right]
\end{cases} \tag{55}
$$

The various estimators being unbiased, the covariance can be expanded as follows:

$$
\begin{aligned}
P_{i|i} &= \text{cov}\left(\theta_i - \hat{\theta}_{i|i} | r_{1:i}\right) \\
&= \text{cov}\left(\theta_i - \left(\hat{\theta}_{i|i-1} + K_i \tilde{r}_i\right) | r_{1:i-1}\right) \\
&= \text{cov}\left(\tilde{\theta}_{i|i-1} - K_i \tilde{r}_i | r_{1:i-1}\right) \\
P_{i|i} &= P_{i|i-1} - P_{\theta r_i} K_i^T - K_i P_{\theta r_i}^T + K_i P_{r_i} K_i^T
\end{aligned} \tag{56}
$$

The optimal gain can thus be obtained by zeroing the gradient with respect to $K_i$ of the trace of this matrix.

First note that the gradient being linear, for three matrices of *ad hoc* dimensions $A$, $B$ and $C$ (that is products $ABA^T$ and $AC^T$ are well defined), $B$ being symmetric, the following algebraic identities hold:

$$
\nabla_A\left(\text{trace}\left(ABA^T\right)\right) = 2AB \tag{57}
$$

$$
\nabla_A\left(\text{trace}\left(AC^T\right)\right) = \nabla_A\left(\text{trace}\left(CA^T\right)\right) = C \tag{58}
$$

and thus using Equation (56) and previous identities:

$$
\begin{aligned}
&\nabla_{K_i}\left(\text{trace}\left(P_{i|i}\right)\right) = 0 \\
\Leftrightarrow \quad & 2K_i P_{r_i} - 2P_{\theta r_i} = 0 \\
\Leftrightarrow \quad & K_i = P_{\theta r_i} P_{r_i}^{-1}
\end{aligned} \tag{59}
$$

Using Equations (56) and (59), the covariance matrix $P_{i|i}$ can be recursively computed as follows:

$$P_{i|i} = P_{i|i-1} - K_i P_{r_i} K_i^T \tag{60}$$

Recall that no Gaussian assumption has been made to derive these equations. Nevertheless, under Gaussian (and linear) assumptions, the optimal update is actually linear[6] (for example, see Chen, 2003). Please also notice that this variance matrix encodes the uncertainty over parameter estimates, and not the intrinsic uncertainty of the considered MDP (it is not the variance of the random process from which the value function is the mean).

### 3.1.3 GENERAL ALGORITHM

The most general KTD algorithm can now be derived. It breaks down in three stages. The first step consists in computing predicted quantities $\hat{\theta}_{i|i-1}$ and $P_{i|i-1}$. These predictions being made from past estimates, the algorithm has to be initialized with priors $\hat{\theta}_{0|0}$ and $P_{0|0}$. Recall that for a random walk model, Equation (50) holds, and the predicted covariance can also be computed analytically:

$$\begin{aligned}
P_{i|i-1} &= \text{cov}\left(\tilde{\theta}_{i|i-1}|r_{1:i-1}\right) \\
&= \text{cov}\left(\tilde{\theta}_{i-1|i-1} + v_i|r_{1:i-1}\right) \\
&= P_{i-1|i-1} + P_{v_i}
\end{aligned} \tag{61}$$

(recall that $P_{v_i}$ is the problem-dependent variance matrix of the evolution noise, to be chosen by the practitioner).

The second step is to compute some statistics of interest. It will be specialized for each algorithm in Section 3.2. The first statistic to compute is the prediction $\hat{r}_{i|i-1}$ (52). The second statistic to compute is the covariance between the parameter vector and the innovation:

$$P_{\theta r_i} = E\left[(\theta_i - \hat{\theta}_{i|i-1})(r_i - \hat{r}_{i|i-1})|r_{1:i-1}\right] \tag{62}$$

However, from the state-space model (46), $r_i = g_{t_i}(\theta_i) + n_i$, and the observation noise is centered and independent, so

$$P_{\theta r_i} = E\left[(\theta_i - \hat{\theta}_{i|i-1})(g_{t_i}(\theta_i) - \hat{r}_{i|i-1})|r_{1:i-1}\right] \tag{63}$$

The last statistic to compute is the covariance of the innovation, which can be written as (using again the characteristics of the observation noise):

$$\begin{aligned}
P_{r_i} &= E\left[(r_i - \hat{r}_{i|i-1})^2|r_{1:i-1}\right] \\
&= E\left[(g_{t_i}(\theta_i) - \hat{r}_{i|i-1} + n_i)^2|r_{1:i-1}\right] \\
&= E\left[(g_{t_i}(\theta_i) - \hat{r}_{i|i-1})^2|r_{1:i-1}\right] + P_{n_i}
\end{aligned} \tag{64}$$

(recall that $P_{n_i}$ is the variance of the observation noise).

---

6. In other words, in this case, the Kalman filtering solution is actually the Bayesian filtering solution.

The third and last step of the algorithm is the correction step. It consists in computing the gain (59), correcting the predicted parameter vector (49) and updating the associated covariance matrix (60) accordingly. The proposed general framework is summarized in Algorithm 1. Notice the similarity between the correction equation ($\hat{\theta}_{i|i} = \hat{\theta}_{i-1|i-1} + K_i(r_i - \hat{r}_{i|i-1})$) and the Widrow-Hoff equation where the approximated value is corrected in the direction of the error (the innovation is indeed the TD error). The gain $K_i$ can be seen as a set of adaptive learning rates.

---

**Algorithm 1**: General KTD algorithm

---

*Initialization*: priors $\hat{\theta}_{0|0}$ and $P_{0|0}$ ;

**for** $i \leftarrow 1, 2, \dots$ **do**

> Observe transition $t_i$ and reward $r_i$ ;
>
> *Prediction step*;
> $\hat{\theta}_{i|i-1} = \hat{\theta}_{i-1|i-1}$;
> $P_{i|i-1} = P_{i-1|i-1} + P_{v_i}$;
>
> *Compute statistics of interest*;
> $\hat{r}_{i|i-1} = E[g_{t_i}(\theta_i)|r_{1:i-1}]$ ;
> $P_{\theta r_i} = E\left[(\theta_i - \hat{\theta}_{i|i-1})(g_{t_i}(\theta_i) - \hat{r}_i)|r_{1:i-1}\right]$;
> $P_{r_i} = E\left[(g_{t_i}(\theta_i) - \hat{r}_{i|i-1})^2|r_{1:i-1}\right] + P_{n_i}$;
>
> *Correction step*;
> $K_i = P_{\theta r_i}P_{r_i}^{-1}$ ;
> $\hat{\theta}_{i|i} = \hat{\theta}_{i|i-1} + K_i\left(r_i - \hat{r}_{i|i-1}\right)$ ;
> $P_{i|i} = P_{i|i-1} - K_iP_{r_i}K_i^T$ ;

---

## 3.2 Specializations

The main difficulty in applying KTD is to compute the statistics of interest $\hat{r}_{i|i-1}$, $P_{\theta r_i}$ and $P_{r_i}$ (for which statistics $\hat{\theta}_{i|i-1}$ and $P_{i|i-1}$ are necessary). First, the value function evaluation in the case of a linear parameterization is considered. The related Bellman equation is (40). In this case an analytical derivation is possible. Then an approximation scheme, the unscented transform (UT) of Julier and Uhlmann (2004), is introduced. It allows solving the same problem for a nonlinear parameterization. $Q$-function evaluation and direct optimization follow.

### 3.2.1 KTD-V: LINEAR PARAMETERIZATION

Here the linear parameterization of equation (8) is adopted, that is $\hat{V}_\theta(s) = \phi(s)^T\theta$. The state-space formulation (46) can thus be rewritten as:

$$\begin{cases} \theta_i = \theta_{i-1} + v_i \\ r_i = \left(\phi(s_i) - \gamma\phi(s_{i+1})\right)^T\theta_i + n_i \end{cases} \tag{65}$$

Notice that as the problem at sight is the evaluation of a deterministic policy, no action has to be observed. The policy being fixed, the MDP reduces to a valued Markov chain. To shorten notations, $H_i$ is defined as:

$$H_i = \phi(s_i) - \gamma\phi(s_{i+1}) \tag{66}$$

As the observation equation is linear, the statistics of interest can be derived analytically. The prediction is:

$$\begin{aligned}
\hat{r}_{i|i-1} &= E\left[g_{t_i}(\theta_i)|r_{1:i-1}\right] \\
&= E\left[H_i^T\theta_i|r_{1:i-1}\right] \\
&= H_i^T E\left[\theta_i|r_{1:i-1}\right] \\
&= H_i^T \hat{\theta}_{i|i-1}
\end{aligned} \tag{67}$$

The covariance between the parameter vector and the innovation can also be computed analytically:

$$\begin{aligned}
P_{\theta r_i} &= E\left[\tilde{\theta}_{i|i-1}\left(g_{t_i}(\theta_i) - \hat{r}_{i|i-1}\right)|r_{1:i-1}\right] \\
&= E\left[\tilde{\theta}_{i|i-1}H_i^T\tilde{\theta}_{i|i-1}|r_{1:i-1}\right] \\
&= E\left[\tilde{\theta}_{i|i-1}\tilde{\theta}_{i|i-1}^T|r_{1:i-1}\right]H_i \\
&= P_{i|i-1}H_i
\end{aligned} \tag{68}$$

The covariance of the innovation is derived analytically as well:

$$\begin{aligned}
P_{r_i} &= E\left[\left(g_{t_i}(\theta_i) - \hat{r}_{i|i-1}\right)^2|r_{1:i-1}\right] + P_{n_i} \\
&= E\left[\left(H_i^T\tilde{\theta}_{i|i-1}\right)^2|r_{1:i-1}\right] + P_{n_i} \\
&= H_i^T P_{i|i-1}H_i + P_{n_i}
\end{aligned} \tag{69}$$

The optimal gain can thus be defined algebraically and recursively:

$$K_i = \frac{P_{i|i-1}H_i}{H_i^T P_{i|i-1}H_i + P_{n_i}} \tag{70}$$

The KTD-V approach for linear parameterization is summarized in Algorithm 2.

Notice that this gain shares similarities with the gain (19) of the LSTD algorithm (Bradtke & Barto, 1996), which is not a surprise. LSTD is based on a least-squares minimization (however with the introduction of instrumental variables in order to handle stochastic transitions), and the Kalman filter can be seen as a stochastic generalization of the least-squares method. This gain shares also similarities with GPTD. Actually, if the process noise is set to 0 (that is $P_{v_i} = 0$), then KTD-V with linear parameterization

---

**Algorithm 2**: KTD-V: linear parameterization

---

*Initialization*: priors $\hat{\theta}_{0|0}$ and $P_{0|0}$ ;

**for** $i \leftarrow 1, 2, \dots$ **do**

    Observe transition $(s_i, s_{i+1})$ and reward $r_i$ ;

    *Prediction step*;
    $\hat{\theta}_{i|i-1} = \hat{\theta}_{i-1|i-1}$;
    $P_{i|i-1} = P_{i-1|i-1} + P_{v_{i-1}}$;

    *Compute statistics of interest*;
    $\hat{r}_{i|i-1} = H_i^T \hat{\theta}_{i|i-1}$ ;
    $P_{\theta r_i} = P_{i|i-1} H_i$;
    $P_{r_i} = H_i^T P_{i|i-1} H_i + P_{n_i}$;
    /*    where $H_i = \phi(s_i) - \gamma \phi(s_{i+1})$                              */

    *Correction step*;
    $K_i = P_{\theta r_i} P_{r_i}^{-1}$ ;
    $\hat{\theta}_{i|i} = \hat{\theta}_{i|i-1} + K_i \left( r_i - \hat{r}_{i|i-1} \right)$ ;
    $P_{i|i} = P_{i|i-1} - K_i P_{r_i} K_i^T$ ;

---

and GPTD are the same algorithm[7], see Equation (22). This is not a surprise: under a linear and Gaussian hypothesis, state-space (65) with zero evolution noise is equivalent to the statistical generative model (21). An alternative point of view is that both approaches provide the least-squares solution to the $L_2$ Bellman residual minimization.

Although linear parameterization is widely used, one can be interested in using a non-linear one (for optimal basis function search or more compact function representation for instance). Another case of interest (addressed later) is to handle the max operator which is inherent to the Bellman optimality equation. This is how the proposed approach notably differs from Engel's framework. Basically, the issue of computing the statistics of interest for KTD can be stated as the following problem: given the mean and covariance of a random variable ($\hat{\theta}_{i|i-1}$ and $P_{i|i-1}$ for KTD), how can the mean and covariance of a nonlinear (and perhaps non-differentiable) mapping ($g_{t_i}$ for KTD) of this random variable be computed? The following section presents the unscented transform, which is an approximation scheme designed to handle such a problem.

---

7. Once again, GPTD is more general than linear parameterization, the gain (22) being refereed to as "parametric GPTD" by Engel (2005). Nevertheless, the non-parametric approach of GPTD actually constructs online a kernel-based linear parameterization. At the end of learning, or if the parameterization is constructed in a preprocessing step, this non-parametric representation reduces to a linear parametric representation. As the focus of this paper is how to learn parameters of a representation and not the representation itself (which we totally recognize as being a problem of importance), GPTD is always considered in its parametric form in this article.

### 3.2.2 The Unscented Transform

Let's abstract from RL and Kalman filtering and consider the problem of non-linear mapping of a random variable. Let $X$ be a random vector, and let $Y$ be a mapping of $X$. The problem is to compute the mean and covariance of $Y$ knowing the mapping and the first and second order moments of $X$. If the mapping is linear, the relation between $X$ and $Y$ can be written as $Y = AX$ where $A$ is a matrix of *ad hoc* dimension (that is number of row of $Y$ times number of rows of $X$). In this case, required mean and covariance can be analytically computed as $E[Y] = AE[X]$ and $E[YY^T] = AE[XX^T]A^T$. This result has been used to derive the KTD-V algorithm of Section 3.2.1.

If the mapping is nonlinear, the relation between $X$ and $Y$ can be written as:

$$Y = f(X) \tag{71}$$

A first solution would be to approximate the nonlinear mapping by a first order Taylor expansion around $E[X]$. This leads to the following approximations of the mean and covariance of $Y$:

$$E[Y] \approx f(E[X]) \tag{72}$$

$$E[YY^T] \approx (\nabla f(E[X])) E[XX^T] (\nabla f(E[X]))^T \tag{73}$$

This approach is the basis of Extended Kalman Filtering (EKF) (for example, see Simon, 2006), which has been extensively studied and used in past decades. However it has some limitations. First it cannot handle non-derivable nonlinearities, and thus cannot handle the Bellman optimality equation (6) because of the max operator. It requires to compute the gradient of the mapping $f$, which can be quite difficult even if possible (eg., neural networks). It also supposes that the nonlinear mapping is locally linearizable in order to have a good approximation, which is unfortunately not always the case and can lead to quite bad results, as exemplified by Julier and Uhlmann (2004).

The basic idea of unscented transform is that it is easier to approximate an arbitrary random vector (with samples) than an arbitrary nonlinear function. Its principle is to sample *deterministically* a set of so-called sigma-points from the expectation and the covariance of $X$. The images of these points through the nonlinear mapping $f$ are then computed, and they are used to approximate statistics of interest. It shares similarities with Monte-Carlo methods, however here the sampling is deterministic and requires less samples to be drawn, nonetheless guaranteeing a given accuracy (Julier & Uhlmann, 2004).

The original unscented transform is now described more formally (some variants have been introduced since then, the basic principle being the same). Let $n$ be the dimension of $X$. A set of $2n + 1$ so-called "sigma-points" is computed as follows:

$$x^{(0)} = \bar{X} \qquad\qquad\qquad j = 0 \tag{74}$$

$$x^{(j)} = \bar{X} + \left(\sqrt{(n+\kappa)P_X}\right)_j \qquad\qquad 1 \le j \le n \tag{75}$$

$$x^{(j)} = \bar{X} - \left(\sqrt{(n+\kappa)P_X}\right)_{j-n} \qquad\qquad n+1 \le j \le 2n \tag{76}$$

as well as associated weights:

$$w_0 = \frac{\kappa}{n+\kappa} \quad \text{and} \quad w_j = \frac{1}{2(n+\kappa)} \quad \forall j > 0 \tag{77}$$

where $\bar{X}$ is the mean of $X$, $P_X$ is its variance matrix, $\kappa$ is a scaling factor which controls the sampling spread, and $(\sqrt{(n+\kappa)P_X})_j$ is the $j^{\text{th}}$ column of the Cholesky decomposition of the matrix $(n+\kappa)P_X$. Then the image through the mapping $f$ is computed for each of these sigma-points:

$$y^{(j)} = f(x^{(j)}), \quad 0 \leq j \leq 2n \tag{78}$$

The set of sigma-points and their images can finally be used to approximate first and second order moments of $Y$, and even $P_{XY}$, the covariance matrix between $X$ and $Y$:

$$\bar{Y} \approx \bar{y} = \sum_{j=0}^{2n} w_j y^{(j)} \tag{79}$$

$$P_Y \approx \sum_{j=0}^{2n} w_j \left( y^{(j)} - \bar{y} \right) \left( y^{(j)} - \bar{y} \right)^T \tag{80}$$

$$P_{XY} \approx \sum_{j=0}^{2n} w_j \left( x^{(j)} - \bar{X} \right) \left( y^{(j)} - \bar{y} \right)^T \tag{81}$$

Thanks to the unscented transform, it is possible to address the value function evaluation problem with nonlinear parameterization, the random vector $X$ being in this case the parameter vector, and its nonlinear mapping $Y$ the predicted reward.

### 3.2.3 KTD-V: NONLINEAR PARAMETERIZATION

In this section a generic parameterization of the value function $\hat{V}_\theta$ is considered: it can be a neural network (Bishop, 1995), a semi-parametric kernel representation (Geist, Pietquin, & Fricout, 2008), or any function representation of interest, as long as it can be described by a set of $p$ parameters. The general state-space formulation (46) can thus be written as:

$$\begin{cases} \theta_i = \theta_{i-1} + v_i \\ r_i = \hat{V}_{\theta_i}(s_i) - \gamma \hat{V}_{\theta_i}(s_{i+1}) + n_i \end{cases} \tag{82}$$

The problem is still to compute the statistics of interest, which becomes tractable with the unscented transform. The first thing to compute is the set of sigma-points from known statistics $\hat{\theta}_{i|i-1}$ and $P_{i|i-1}$ as well as the associated weights using Equations (74-77), as described in Section 3.2.2:

$$\Theta_{i|i-1} = \left\{ \hat{\theta}_{i|i-1}^{(j)}, \ 0 \leq j \leq 2p \right\} \tag{83}$$

$$\mathcal{W} = \{w_j, \ 0 \leq j \leq 2p\} \tag{84}$$

Then the images of these sigma-points are computed (a predicted reward for each of the sampled parameter vectors), using the observation function of state-space model (82), which is linked to the Bellman evaluation equation (40):

$$\mathcal{R}_{i|i-1} = \left\{ \hat{r}_{i|i-1}^{(j)} = \hat{V}_{\hat{\theta}_{i|i-1}^{(j)}}(s_i) - \gamma \hat{V}_{\hat{\theta}_{i|i-1}^{(j)}}(s_{i+1}), \ 0 \leq j \leq 2p \right\} \tag{85}$$

The sigma-points and their images being computed, the statistics of interest can be approximated by:

$$\hat{r}_{i|i-1} \approx \sum_{j=0}^{2p} w_j \hat{r}_{i|i-1}^{(j)} \tag{86}$$

$$P_{r_i} \approx \sum_{j=0}^{2p} w_j \left( \hat{r}_{i|i-1}^{(j)} - \hat{r}_{i|i-1} \right)^2 + P_{n_i} \tag{87}$$

$$P_{\theta r_i} \approx \sum_{j=0}^{2p} w_j \left( \hat{\theta}_{i|i-1}^{(j)} - \hat{\theta}_{i|i-1} \right) \left( \hat{r}_{i|i-1}^{(j)} - \hat{r}_{i|i-1} \right) \tag{88}$$

As the unscented transform is no longer an approximation for linear mapping, this formulation is still valid for value function evaluation with linear function approximation. KTD-V with nonlinear function approximation is summarized in Algorithm 3. Notice that such a general parameterization cannot be taken into account in GPTD nor LSTD. It is possible with direct algorithms (TD with function approximation), however there is a risk of divergence. This is illustrated in Section 7.

### 3.2.4 KTD-SARSA

This section focuses on the $Q$-function evaluation of a fixed given policy. The associated algorithm is called KTD-SARSA, which can be misleading. Indeed, SARSA is sometime understood as a $Q$-function evaluation algorithm associated with an optimistic policy iteration scheme (eg., $\epsilon$-greedy policy). Here the focus is on the $Q$-function evaluation problem, and the control part is left apart. For a general parameterization $\hat{Q}_\theta$, and considering the Bellman evaluation equation (41), the state-space model (46) can be rewritten as:

$$\begin{cases} \theta_i = \theta_{i-1} + v_i \\ r_i = \hat{Q}_{\theta_i}(s_i, a_i) - \gamma \hat{Q}_{\theta_i}(s_{i+1}, a_{i+1}) + n_i \end{cases} \tag{89}$$

For a fixed policy, the value function evaluation on the state space induced Markov chain[8] is quite similar to the $Q$-function evaluation on the state-action space induced Markov chain. It is thus straightforward to extend KTD-V to $Q$-function evaluation. Recall that for a linear parameterization, the unscented transform leads to an exact computation of statistics of interest, and thus in this case Algorithm 3 (KTD-V) is equivalent to Algorithm 2. That is why only the sigma-point formulation of KTD-SARSA is given, also summarized in Algorithm 3.

LSTD and GPTD have also been generalized to the $Q$-function evaluation (see respectively Lagoudakis & Parr, 2003 and Engel, 2005). However, once again, these approaches cannot handle a nonlinear parameterization, contrary to KTD-SARSA. Notice also that if the parameterization is linear and the process noise is zero, KTD-SARSA is the same algorithm as GPTD for $Q$-function evaluation (this is a direct extension of the equivalence between GPTD and KTD-V with linear parameterization and zero process noise, see Sec. 3.2.1).

---

8. For a fixed policy, the MDP reduces to a Markov chain.

---

**Algorithm 3**: KTD-V, KTD-SARSA and KTD-Q

---

*Initialization*: priors $\hat{\theta}_{0|0}$ and $P_{0|0}$ ;

**for** $i \leftarrow 1, 2, \dots$ **do**

Observe transition $t_i = \begin{cases} (s_i, s_{i+1}) \text{ (KTD-V)} \\ (s_i, a_i, s_{i+1}, a_{i+1}) \text{ (KTD-SARSA)} \\ (s_i, a_i, s_{i+1}) \text{ (KTD-Q)} \end{cases}$ and reward $r_i$ ;

*Prediction Step*;
$\hat{\theta}_{i|i-1} = \hat{\theta}_{i-1|i-1}$;
$P_{i|i-1} = P_{i-1|i-1} + P_{v_i}$;

*Sigma-points computation* ;
$\Theta_{i|i-1} = \left\{ \hat{\theta}_{i|i-1}^{(j)}, \quad 0 \leq j \leq 2p \right\}$ (from $\hat{\theta}_{i|i-1}$ and $P_{i|i-1}$);
$\mathcal{W} = \{ w_j, \quad 0 \leq j \leq 2p \quad \}$ ;
$\mathcal{R}_{i|i-1} =$
$\begin{cases} \left\{ \hat{r}_{i|i-1}^{(j)} = \hat{V}_{\hat{\theta}_{i|i-1}^{(j)}}(s_i) - \gamma \hat{V}_{\hat{\theta}_{i|i-1}^{(j)}}(s_{i+1}), 0 \leq j \leq 2p \right\} \text{ (KTD-V)} \\ \left\{ \hat{r}_{i|i-1}^{(j)} = \hat{Q}_{\hat{\theta}_{i|i-1}^{(j)}}(s_i, a_i) - \gamma \hat{Q}_{\hat{\theta}_{i|i-1}^{(j)}}(s_{i+1}, a_{i+1}), 0 \leq j \leq 2p \right\} \text{ (KTD-SARSA)} \\ \left\{ \hat{r}_{i|i-1}^{(j)} = \hat{Q}_{\hat{\theta}_{i|i-1}^{(j)}}(s_i, a_i) - \gamma \max_{b \in A} \hat{Q}_{\hat{\theta}_{i|i-1}^{(j)}}(s_{i+1}, b), 0 \leq j \leq 2p \right\} \text{ (KTD-Q)} \end{cases}$ ;

*Compute statistics of interest*;
$\hat{r}_{i|i-1} = \sum_{j=0}^{2p} w_j \hat{r}_{i|i-1}^{(j)}$;
$P_{\theta r_i} = \sum_{j=0}^{2p} w_j (\hat{\theta}_{i|i-1}^{(j)} - \hat{\theta}_{i|i-1})(\hat{r}_{i|i-1}^{(j)} - \hat{r}_{i|i-1})$;
$P_{r_i} = \sum_{j=0}^{2p} w_j \left( \hat{r}_{i|i-1}^{(j)} - \hat{r}_{i|i-1} \right)^2 + P_{n_i}$;

*Correction step*;
$K_i = P_{\theta r_i} P_{r_i}^{-1}$ ;
$\hat{\theta}_{i|i} = \hat{\theta}_{i|i-1} + K_i \left( r_i - \hat{r}_{i|i-1} \right)$ ;
$P_{i|i} = P_{i|i-1} - K_i P_{r_i} K_i^T$ ;

---

### 3.2.5 KTD-Q

This section focuses on the $Q$-function optimization, that is on finding an approximate solution to the Bellman optimality equation (42). A general parameterization $\hat{Q}_\theta$ is adopted. The state-space model (46) can be specialized as follows:

$$\begin{cases} \theta_i = \theta_{i-1} + v_i \\ r_i = \hat{Q}_{\theta_i}(s_i, a_i) - \gamma \max_{b \in A} \hat{Q}_{\theta_i}(s_{i+1}, b) + n_i \end{cases} \tag{90}$$

Here linear and nonlinear parameterizations are not distinguished, because of the nonlinearities induced by the max operator. It is tricky to handle, especially because of its non-differentiability.

Hopefully, as it approximates the random variable rather than the mapping, the unscented transform is a derivative-free approximation. Given the general KTD algorithm introduced in Section 3.1.3 and the unscented transform described in Section 3.2.2, it is possible to derive KTD-Q, the KTD algorithm for $Q$-function direct optimization. One has first to compute the set of sigma-points associated with the parameter vector, as in equations (83-84). Then the mapping of these sigma-points through the observation equation of state-space model (90), which contains the max operator, is computed:

$$\mathcal{R}_{i|i-1} = \left\{ \hat{r}_{i|i-1}^{(j)} = \hat{Q}_{\hat{\theta}_{i|i-1}^{(j)}}(s_i, a_i) - \gamma \max_{b \in A} \hat{Q}_{\hat{\theta}_{i|i-1}^{(j)}}(s_{i+1}, b), \ 0 \le j \le 2p \right\} \quad (91)$$

Then, as usual, the sigma-points and their images are used to compute the statistics of interest, as in equations (86-88). The proposed KTD-Q is summarized in Algorithm 3.

Notice that even if the parameterization is linear, there is no LSTD nor GPTD equivalent to this algorithm. Actually, as linearity of the observation model is a mandatory assumption for the derivation of these algorithms, the Bellman optimality operator cannot be taken into account. As far as we know, KTD-Q is one of the first second order value iteration-like algorithms. Choi and Van Roy (2006) propose a linear least-squares based bootstrapping approach (to be discussed in Section 8) which can be used in a Q-learning-like setting. Yu and Bertsekas (2007) also introduce a least-squares-based Q-learning. However, it is designed for optimal stopping problems (which is a restrictive class of MDP) and it is not truly online (to update the representation given a new observation, all the followed trajectory are explicitly required). Roughly speaking, this algorithm is fitted-Q with a least-squares for the supervised learning part and for which a new transition is added to the learning basis at each iteration. Its computational complexity is cubic[9], which is higher than the square complexity of KTD, as shown in the next section.

## 3.3 Algorithmic Complexity

Let $p$ be the number of parameters. The unscented transform involves a Cholesky decomposition of which computational complexity is $O(p^3)$ in general. However, as the variance update (60) is a rank one update, the Cholesky decomposition can be perfomed in $O(p^2)$ (eg., see Gill, Golub, Murray, & Saunders, 1974). The different algorithms imply to evaluate $2p + 1$ times the $g_{t_i}$ function at each time-step. For KTD-V or KTD-SARSA and a general parameterization, each evaluation is bounded by $O(p)$. For KTD-Q, the maximum over actions has to be computed. The notation $\mathcal{A}$ represents the cardinality of action space if finite, the computational complexity of the algorithm used to search the maximum otherwise (eg., the number of samples times the evaluation complexity for Monte Carlo). Then each evaluation is bounded by $O(p\mathcal{A})$. Remaining operations are basic linear algebra, and are thus bounded by $O(p^2)$. Therefore the global computational complexity (per iteration) of KTD-V and KTD-SARSA is $O(p^2)$, and KTD-Q is in $O(\mathcal{A}p^2)$. As the mean and variance matrix of parameters have to be maintained, the memory complexity is $O(p^2)$. Although comparable to LSTD or GPTD complexity, this is higher than many other RL algorithms which have a linear complexity. Nevertheless, most of value function approximation approaches assume a linear parameterization. KTD does not make this hypothesis (even to

9. However, the paper proposes some heuristics which reduce this complexity.

analyse convergence, as shown in Section 5.1) and so allows much more compact representations for the value function. Thus the quadratic complexity is a problem with important counterparts.

## 4. KTD: the Stochastic Case

The KTD framework presented so far assumes deterministic transitions. If it is not the case, the observation noise $n_i$ cannot be assumed as white (since it would include the MDP stochasticity as well as the inductive bias), whereas it is a necessary condition for KTD derivation. First it is shown that using KTD in a stochastic MDP involves a bias. Then a colored noise model is introduced to alleviate this problem, and it is used to extend KTD. The problem caused by off-policy learning, which prevents the derivation of an XKTD-Q algorithm, is also discussed.

### 4.1 Stochastic Transitions and Bias

One can ignore this problem and use the cost function (47) linked to state-space model (46) with stochastic transitions. However, similarly to approaches minimizing a squared Bellman residual, such as residual algorithms of Baird (1995), this cost function is biased. More precisely, it is biased relatively to stochasticity of transitions (parameters and transitions are different sources of randomness). Additionally, this cost function being biased, the estimator minimizing it (that is $\hat{\theta}_{i|i}$) is biased too.

**Theorem 1.** *If the reward function only depends on the current state-action pair, and not on the transiting state, then when used on a stochastic Markov decision process, the cost function (47) is biased (relatively to stochasticity of transitions), its bias being given by:*

$$\|K_i\|^2 E\left[\operatorname*{cov}_{s'|s_i,a_i}(r_i - g_{t_i}(\theta))\,|r_{1:i-1}\right] = \begin{cases} \|K_i\|^2 E\left[\operatorname{cov}_{s'|s_i,\pi(s_i)}(r_i + \gamma V_\theta(s'))\,|r_{1:i-1}\right] \\ \|K_i\|^2 E\left[\operatorname{cov}_{s'|s_i,\pi(s_i)}(r_i + \gamma Q_\theta(s',\pi(s')))\,|r_{1:i-1}\right] \\ \|K_i\|^2 E\left[\operatorname{cov}_{s'|s_i,a_i}(r_i + \gamma \max_{a\in A} Q_\theta(s',a))\,|r_{1:i-1}\right] \end{cases}$$

(92)

*It is clear that this bias is zero for deterministic transitions.*

*Proof.* The assumption that the reward does not depend on the transiting state is made for technically simplifying the demonstration, because of the conditioning of the cost function on past observed rewards. Yet it is done without loss of generality. Under this hypothesis, the state-space model to be considered for a stochastic MDP is:

$$\begin{cases} \theta_i = \theta_{i-1} + v_i \\ r_i = E_{s'|s_i,a_i}[g_{t_i}(\theta_i)] + n_i \end{cases}$$

(93)

with $t_i$ now defined as the random quantity $t_i = (s_i, a_i, s')$. Notice that the observation equation (minus the noise) is the Bellman equation for stochastic transitions. The difference with state-space model (46) is that transitions are no more sampled but averaged. The associated cost function is:

$$\mathcal{J}_i(\theta) = \operatorname{trace}\left(\mathcal{P}_{i|i}\right) = \operatorname{trace}\left(\mathcal{P}_{i|i-1} - \mathcal{P}_{\theta r_i} K_i^T - K_i \mathcal{P}_{\theta r_i}^T - K_i \mathcal{P}_{r_i} K_i^T\right)$$

(94)

Calligraphic letters denote the same for state-space model (93) than notations (55) for state-space model (46), eg.:

$$\mathcal{P}_{\theta r_i} = E\left[\tilde{\theta}_{i|i-1}\tilde{\mathfrak{r}}_i|r_{1:i-1}\right] \text{ with } \tilde{\mathfrak{r}}_i = r_i - \hat{\mathfrak{r}}_{i|i-1} = r_i - E\left[E_{s'|s_i,a_i}\left[g_{t_i}(\theta_i)\right]|r_{1:i-1}\right] \quad (95)$$

Notice that the prediction of the reward is unbiased, thus the same holds for the innovation:

$$E_{s'|s_i,a_i}\left[\hat{r}_{i|i-1}\right] = \hat{\mathfrak{r}}_{i|i-1} \text{ and } E_{s'|s_i,a_i}\left[\tilde{r}_{i|i-1}\right] = \tilde{\mathfrak{r}}_{i|i-1} \quad (96)$$

The term $P_{i|i-1}$ does not depend on transiting state $s'$ and the term $P_{\theta r_i}$ is linear in the innovation, so they are unbiased:

$$E_{s'|s_i,a_i}\left[P_{i|i-1}\right] = \mathcal{P}_{i|i-1} \text{ and } E_{s'|s_i,a_i}\left[P_{\theta r_i}\right] = \mathcal{P}_{\theta r_i} \quad (97)$$

This is not the case for the variance of the innovation:

$$\begin{aligned}
E_{s'|s_i,a_i}\left[P_{r_i}\right] &= E_{s'|s_i,a_i}\left[E\left[\tilde{r}_i^2|r_{1:i-1}\right]\right] \\
&= E\left[E_{s'|s_i,a_i}\left[\tilde{r}_i^2\right]|r_{1:i-1}\right] \\
&= E\left[\tilde{\mathfrak{r}}_i^2|r_{1:i-1}\right] + E\left[E_{s'|s_i,a_i}\left[\tilde{r}_i^2\right] - \left(E_{s'|s_i,a_i}\left[\tilde{r}_i\right]\right)^2|r_{1:i-1}\right] \\
&= \mathcal{P}_{r_i} + E\left[\operatorname*{cov}_{s'|s_i,ai}(\tilde{r}_i)|r_{1:i-1}\right]
\end{aligned} \quad (98)$$

Thus the bias $(E_{s'|s_i,a_i}[J_i(\theta)] - \mathcal{J}_i(\theta))$ can be computed:

$$\begin{aligned}
E_{s'|s_i,a_i}\left[J_i(\theta)\right] - \mathcal{J}_i(\theta) &= E_{s'|s_i,a_i}\left[\operatorname{trace}\left(K_i\left(P_{r_i} - \mathcal{P}_{r_i}\right)K_i^T\right)\right] \\
&= \operatorname{trace}(K_iK_i^T)E_{s'|s_i,a_i}\left[P_{r_i} - \mathcal{P}_{r_i}\right] \\
&= K_i^T K_i\left(E_{s'|s_i,a_i}\left[P_{r_i}\right] - \mathcal{P}_{r_i}\right) \\
&= \|K_i\|^2 E\left[\operatorname*{cov}_{s'|s_i,a_i}(r_i - g_{t_i}(\theta))|r_{1:i-1}\right]
\end{aligned} \quad (99)$$

Notice that neither $V_\theta(s_i)$ nor $Q_\theta(s_i,a_i)$ depends on the transiting state $s'$. Thus this proves the result as expressed in Theorem 1. □

This bias is quite similar to the one arising from the minimization of a square Bellman residual. The result of Theorem 2 (see Section 5) even strengthen this parallel. A solution could be to introduce an auxiliary filter to remove this bias, similarly to introduction of an auxiliary function made by Antos, Szepesvári, and Munos (2008). However extension of this work is not straightforward. Another approach could be to estimate this bias online so as to remove it, similarly to what is done by Jo and Kim (2005) for least-mean square filtering. However the Kalman filter is a much more complex framework than the least-squares filter, especially when combined with unscented transform. Another interesting perspective could be to introduce a colored observation noise as done by Engel (2005) in a Bayesian context for Gaussian process-based algorithms. This last approach is presented and used to extend KTD next.

## 4.2 A Colored Noise Model

First the focus is on value function evaluation. Extension to $Q$-function evaluation is straightforward, and $Q$-function optimization is discussed later, because of its *off-policy* aspect (the learnt policy is not the behaviorial one). The Bellman evaluation equation to be solved is Equation (4): it has just been shown that directly using KTD in a stochastic problem induces a bias in the minimized cost function. A colored noise model which was first proposed by Engel et al. (2005) (the basis of the so-called Monte-Carlo GPTD algorithm) is first presented, before being adapted to extend the KTD framework.

The policy being fixed for evaluation, the MDP reduces in a valued Markov chain of probability transition $p^\pi(.|s) = p(.|s, \pi(s))$ and of reward $R^\pi(s, s') = R(s, \pi(s), s')$. The value function can be defined as the expectation (over all possible trajectories) of the following discount return random process:

$$D^\pi(s) = \sum_{i=0}^{\infty} \gamma^i R^\pi(s_i, s_{i+1}) | s_0 = s, \ s_{i+1} \sim p^\pi(.|s_i) \tag{100}$$

This equation naturally leads to a Bellman-like anti-causal recurrence:

$$D^\pi(s) = R^\pi(s, s') + \gamma D^\pi(s'), \ s' \sim p^\pi(.|s) \tag{101}$$

This random process can also be broken down in its mean plus a zero mean residual. However by definition its mean is the value function $V^\pi(s) = E[D^\pi(s)]$, so by writing $\Delta V^\pi(s)$ the residual:

$$D^\pi(s) = E[D^\pi(s)] + (D^\pi(s) - E[D^\pi(s)]) = V^\pi(s) + \Delta V^\pi(s) \tag{102}$$

Substituting Equation (102) into Equation (101), the reward can be expressed as a function of the value plus a noise:

$$R^\pi(s, s') = V^\pi(s) - \gamma V^\pi(s') + N(s, s') \tag{103}$$

the noise being defined as:

$$N(s, s') = \Delta V^\pi(s) - \gamma \Delta V^\pi(s') \tag{104}$$

As done by Engel et al. (2005), the residuals are supposed to be independent, which leads to a colored noise model. This assumption is really strong, as transitions are likely to render residuals dependent, however despite this some convergence guarantees are given in Section 5.

Recall the observation equation of the state-space formulation (46): $r_i = g_{t_i}(\theta_i) + n_i$. In the KTD framework, the observation noise $n_i$ is assumed white, which is necessary for the algorithm derivation. In the eXtended Kalman Temporal Differences (XKTD) framework, the colored noise model (104) is used instead.

The residual being centered and assumed independent, this noise is indeed a moving average (MA) noise (here the sum of two white noises):

$$n_i = -\gamma u_i + u_{i-1}, \quad u_i \sim (0, \sigma_i^2) \tag{105}$$

Notice that the white noise $u_i$ is centered with variance $\sigma_i^2$, nevertheless no assumption is made about its distribution (particularly no Gaussian assumption).

### 4.3 Extending KTD

It is quite easy to use an autoregressive (AR) process noise in a Kalman filter by extending the evolution equation (for example, see Simon, 2006). However, as far as we know, the case of an MA observation noise has never been addressed before in the literature, whereas it is necessary to extend KTD. Notice that this noise model is taken into account in a quite different way in the GPTD framework. Basically, it is done using the partitioned matrix inversion formula, which is not possible here due to the lack of linearity assumption.

#### 4.3.1 EXTENDED KALMAN TEMPORAL DIFFERENCES

Rederiving KTD in the case of an MA noise as done in Section 3.1 would be quite difficult. Instead, it is proposed here to express the scalar MA noise $n_i$ as a vectorial AR noise. This allows extending state-space model (46) to a new one for which Algorithm 1 applies rather directly. Let $\omega_i$ be an auxiliary random variable. Scalar MA noise (105) is equivalent to the following vectorial AR noise:

$$\begin{pmatrix} \omega_i \\ n_i \end{pmatrix} = \begin{pmatrix} 0 & 0 \\ 1 & 0 \end{pmatrix} \begin{pmatrix} \omega_{i-1} \\ n_{i-1} \end{pmatrix} + \begin{pmatrix} 1 \\ -\gamma \end{pmatrix} u_i \tag{106}$$

Indeed, from this vectorial AR noise, $n_i = \omega_{i-1} - \gamma u_i$ and $\omega_i = u_i$, so $n_i = -\gamma u_i + u_{i-1}$ which is the correct MA model. The noise $u_i' = \begin{pmatrix} u_i & -\gamma u_i \end{pmatrix}^T$ is also centered and its variance matrix is:

$$P_{u_i'} = \sigma_i^2 \begin{pmatrix} 1 & -\gamma \\ -\gamma & \gamma^2 \end{pmatrix} \tag{107}$$

This new noise formulation having been defined, it is now possible to extend the state-space formulation (46):

$$\begin{cases} \mathbf{x}_i = F\mathbf{x}_{i-1} + v_i' \\ r_i = g_{t_i}(\mathbf{x}_i) \end{cases} \tag{108}$$

The parameter vector is now extended with the vectorial AR noise $\begin{pmatrix} \omega_i & n_i \end{pmatrix}^T$:

$$\mathbf{x}_i^T = \begin{pmatrix} \theta_i^T & \omega_i & n_i \end{pmatrix} \tag{109}$$

Notice that as the observation noise $n_i$ is now a part of the extended parameter vector, it is also estimated. The evolution matrix $F$ takes into account the structure of the MA observation noise. Let $p$ be the number of parameters and $I_p$ the identity matrix of size p, the evolution matrix is written by bloc ($\mathbf{0}$ denotes a zero $p \times 1$ column vector):

$$F = \begin{pmatrix} I_p & \mathbf{0} & \mathbf{0} \\ \mathbf{0}^T & 0 & 0 \\ \mathbf{0}^T & 1 & 0 \end{pmatrix} \tag{110}$$

The process noise $v_i$ is also extended to take into account the MA observation noise. It is still centered, however its variance matrix is extended using the variance matrix $P_{u_i'}$ (107):

$$P_{v_i'} = \begin{pmatrix} P_{v_i} & \mathbf{0} & \mathbf{0} \\ \mathbf{0}^T & \sigma_i^2 & -\gamma\sigma_i^2 \\ \mathbf{0}^T & -\gamma\sigma_i^2 & \gamma^2\sigma_i^2 \end{pmatrix} \tag{111}$$

The observation equation remains the same:

$$r_i = g_{t_i}(\mathbf{x}_i) = g_{t_i}(\theta_i) + n_i \tag{112}$$

However now the observation noise is a part of the evolution equation, and it has to be estimated.

Using this new state-space formulation, a general XKTD algorithm can be derived. It is summarized in Algorithm 4. It is rather similar to Algorithm 1 with two slight changes: the state-space to be considered is now given by Equation (108) and prediction of mean and covariance of the extended random vector $\mathbf{x}_i$ is done using the evolution matrix $F$ (which is the identity for KTD). Notice that the computational complexity is the same for both algorithms, as the parameter vector is extended with only two scalars. As for KTD, XKTD can be specialized to XKTD-V (value function evaluation) and XKTD-SARSA ($Q$-function evaluation). The reasoning is the same as in Section 3.2 and practical approaches are given in Algorithm 5. Yet, specialization to XKTD-Q is not straightforward because of its off-policy nature, as explained in section 4.3.2.

Recall that KTD with zero process noise and linear parameterization is the same algorithm as GPTD (see Sec. 3.2.1). Actually, the same holds for XKTD with zero process noise and linear parameterization and MC-GPTD (the algorithm obtained using the same colored noise model in the GPTD framework, however in a different manner, see Engel et al., 2005). This can be easily (but lengthly) checked by expanding XKTD equations in the linear case. Once again, MC-GPTD can certainly be extended to handle non-stationarities, even if it is less natural than for XKTD, but it cannot handle nonlinear parameterization. From this point of view, XKTD extends MC-GPTD.

---

**Algorithm 4**: General XKTD algorithm

---

*Initialization*: priors $\hat{\mathbf{x}}_{0|0}$ and $P_{0|0}$ ;

**for** $i \leftarrow 1, 2, \ldots$ **do**

    Observe transition $t_i$ and reward $r_i$ ;

    *Prediction step*;
    $\hat{\mathbf{x}}_{i|i-1} = F\hat{\mathbf{x}}_{i-1|i-1}$;
    $P_{i|i-1} = FP_{i-1|i-1}F^T + P_{v_i'}$;

    *Compute statistics of interest (using UT)*;
    $\hat{r}_{i|i-1} = E[g_{t_i}(\theta_i) + n_i | r_{1:i-1}]$ ;
    $P_{\mathbf{x}r_i} = E\left[(\mathbf{x}_i - \hat{\mathbf{x}}_{i|i-1})(g_{t_i}(\theta_i) + n_i - \hat{r}_{i|i-1}) | r_{1:i-1}\right]$;
    $P_{r_i} = E\left[(g_{t_i}(\theta_i) + n_i - \hat{r}_{i|i-1})^2 | r_{1:i-1}\right]$;

    *Correction step*;
    $K_i = P_{\mathbf{x}r_i}P_{r_i}^{-1}$ ;
    $\hat{\mathbf{x}}_{i|i} = \hat{\mathbf{x}}_{i|i-1} + K_i\left(r_i - \hat{r}_{i|i-1}\right)$ ;
    $P_{i|i} = P_{i|i-1} - K_iP_{r_i}K_i^T$ ;

---

---

**Algorithm 5**: XKTD-V and XKTD-SARSA

---

*Initialization*: priors $\hat{\mathbf{x}}_{0|0} = \left(\hat{\theta}_{0|0}^T \quad 0 \quad 0\right)^T$ and $P_{0|0}$ ;

**for** $i \leftarrow 1, 2, \dots$ **do**

Observe transition $t_i = \begin{cases} (s_i, s_{i+1}) \text{ (XKTD-V)} \\ (s_i, a_i, s_{i+1}, a_{i+1}) \text{ (XKTD-SARSA)} \end{cases}$ and reward $r_i$ ;

*Prediction Step*;
$\hat{\mathbf{x}}_{i|i-1} = F\hat{\mathbf{x}}_{i-1|i-1}$;
$P_{i|i-1} = FP_{i-1|i-1}F^T + P_{v'_i}$;

*Sigma-points computation* ;
$\mathbf{X}_{i|i-1} = \left\{\hat{\mathbf{x}}_{i|i-1}^{(j)}, \quad 0 \leq j \leq 2p+4\right\}$ (from $\hat{\mathbf{x}}_{i|i-1}$ and $P_{i|i-1}$);
$\mathcal{W} = \{w_j, \quad 0 \leq j \leq 2p+4 \ \}$ ;
/* notice that $(\hat{\mathbf{x}}_{i|i-1}^{(j)})^T = \left((\hat{\theta}_{i|i-1}^{(j)})^T \quad \hat{\omega}_{i|i-1}^{(j)} \quad \hat{n}_{i|i-1}^{(j)}\right)$ */
$\mathcal{R}_{i|i-1} =$
$\begin{cases} \left\{\hat{r}_{i|i-1}^{(j)} = \hat{V}_{\hat{\theta}_{i|i-1}^{(j)}}(s_i) - \gamma \hat{V}_{\hat{\theta}_{i|i-1}^{(j)}}(s_{i+1}) + \hat{n}_{i|i-1}^{(j)}, 0 \leq j \leq 2p+4\right\} \text{ (XKTD-V)} \\ \left\{\hat{r}_{i|i-1}^{(j)} = \hat{Q}_{\hat{\theta}_{i|i-1}^{(j)}}(s_i, a_i) - \gamma \hat{Q}_{\hat{\theta}_{i|i-1}^{(j)}}(s_{i+1}, a_{i+1}) + \hat{n}_{i|i-1}^{(j)}, 0 \leq j \leq 2p+4\right\} \text{ (XKTD-SARSA)} \end{cases}$ ;

*Compute statistics of interest*;
$\hat{r}_{i|i-1} = \sum_{j=0}^{2p+4} w_j \hat{r}_{i|i-1}^{(j)}$;
$P_{\mathbf{x}r_i} = \sum_{j=0}^{2p+4} w_j (\hat{\mathbf{x}}_{i|i-1}^{(j)} - \hat{\mathbf{x}}_{i|i-1})(\hat{r}_{i|i-1}^{(j)} - \hat{r}_{i|i-1})$;
$P_{r_i} = \sum_{j=0}^{2p+4} w_j \left(\hat{r}_{i|i-1}^{(j)} - \hat{r}_{i|i-1}\right)^2$;

*Correction step*;
$K_i = P_{\mathbf{x}r_i} P_{r_i}^{-1}$ ;
$\hat{\mathbf{x}}_{i|i} = \hat{\mathbf{x}}_{i|i-1} + K_i \left(r_i - \hat{r}_{i|i-1}\right)$ ;
$P_{i|i} = P_{i|i-1} - K_i P_{r_i} K_i^T$ ;

---

### 4.3.2 XKTD and Off-policy Learning

Off-policy learning is the problem of learning the value of one policy (the target policy) while following another one (the behavior policy). KTD-Q (or more generally $Q$-learning-like algorithms) is an example of off-policy learning: the behavior policy is any sufficiently exploratory policy while the learnt policy is the optimal one. More generally, off-policy learning is of interest, for example to reuse previous trajectories or if the behavioral policy cannot be controlled.

Using a colored observation noise results in a memory effect, similarly to what happens with eligibility traces for more classical TD algorithms (Sutton & Barto, 1998). As classical eligibility-trace algorithms, XKTD applied to off-policy learning should fail because it includes some effect of multi-step transitions, which are contaminated by the behavior policy and not compensated for in any way. For a discussion about off-policy learning and memory effects, see for example the work of Precup, Sutton, and Singh (2000). The link of this memory effect to Monte Carlo (and to eligibility traces when the eligibility factor is set to 1) is shown in the convergence analysis of Section 5. Here it is analyzed through XKTD equations by showing that parameters are updated according to all past temporal differences errors, and not only the current one.

To show this, a first step is to expand the prediction equation:

$$\hat{\mathbf{x}}_{i|i-1} = F\hat{\mathbf{x}}_{i-1|i-1}$$
$$\Leftrightarrow \begin{pmatrix} \hat{\theta}_{i|i-1} \\ \hat{\omega}_{i|i-1} \\ \hat{n}_{i|i-1} \end{pmatrix} = \begin{pmatrix} \hat{\theta}_{i-1|i-1} \\ 0 \\ \hat{\omega}_{i-1|i-1} \end{pmatrix} \tag{113}$$

Let $\hat{g}_{t_i}$ be defined as:

$$\hat{g}_{t_i} = E[g_{t_i}(\theta_i)|r_{1:i-1}] \tag{114}$$

In the KTD framework, $\hat{g}_{t_i}$ is actually the predicted reward. However, it is not the case in the XKTD framework, because the estimated noise has also to be taken into account. The predicted reward can be expanded using Eq. (113):

$$\hat{r}_{i|i-1} = E[g_{t_i}(\theta_i) + n_i|r_{1:i-1}]$$
$$= \hat{g}_{t_i} + \hat{n}_{i|i-1}$$
$$= \hat{g}_{t_i} + \hat{\omega}_{i-1|i-1} \tag{115}$$

A blockwise notation is adopted for the Kalman gain:

$$K_i = \begin{pmatrix} K_{\theta_i} \\ K_{\omega_i} \\ K_{n_i} \end{pmatrix} \tag{116}$$

This being stated, the correction equation can be expanded:

$$\hat{\mathbf{x}}_{i|i} = \hat{\mathbf{x}}_{i|i-1} + K_i\tilde{r}_i$$
$$\Leftrightarrow \begin{pmatrix} \hat{\theta}_{i|i} \\ \hat{\omega}_{i|i} \\ \hat{n}_{i|i} \end{pmatrix} = \begin{pmatrix} \hat{\theta}_{i-1|i-1} \\ 0 \\ \hat{\omega}_{i-1|i-1} \end{pmatrix} + \begin{pmatrix} K_{\theta_i} \\ K_{\omega_i} \\ K_{n_i} \end{pmatrix} \left( r_i - \hat{g}_{t_i} - \hat{\omega}_{i-1|i-1} \right) \tag{117}$$

From the last equation a general update of the parameters can be derived:

$$\hat{\theta}_{i|i} = \hat{\theta}_{i-1|i-1} + K_{\theta_i} \left( r_i - \hat{g}_{t_i} - K_{w_{i-1}}\tilde{r}_{i-1} \right) \tag{118}$$

The parameters are thus updated according to the temporal difference error at time $i$, $\delta_i = r_i - \hat{g}_{t_i}$, and to the innovation at time $i-1$, $\tilde{r}_{i-1}$, which is itself (by recurrence) a combination of TD error at time $i-1$ and of innovation at time $i-2$, *etc.* This update equation highlights the memory effect of XKTD which prevents its use in an off-policy learning scenario. Notably, this prevents the derivation of a XKTD-Q algorithm. A solution to combine off-policy learning and the colored noise could be to use some importance sampling scheme, a well known approach of the Monte Carlo literature which allows estimating quantities linked to a distribution using samples drawn from another distribution.

## 5. Convergence Analysis

This section provides a convergence analysis for both KTD (deterministic MDPs) and XKTD (stochastic MDPs).

### 5.1 Deterministic Case

First a convergence analysis of the KTD algorithm is provided for deterministic MDP. It leads to a result similar to the one of residual algorithms (Baird, 1995), that is the minimization of the squared Bellman residual. This theorem makes some strong assumptions (actually the same as the GPTD framework, however without the linear hypothesis). However, it is important to remark that even if these hypotheses are not satisfied, the cost function (47) is still minimized. The aim of this result is to link KTD to more classic RL algorithms.

**Theorem 2.** *Under the assumptions that posterior and noise distributions are Gaussian and that the prior is Gaussian too (of mean $\theta_0$ and variance $P_0$), than the Kalman Temporal Differences algorithm (white observation noise assumption) minimizes the following regularized empirical cost function:*

$$C_i(\theta) = \sum_{j=0}^{i} \frac{1}{P_{n_j}} \left( r_j - g_{t_j}(\theta) \right)^2 + (\theta - \theta_0)^T P_0^{-1} (\theta - \theta_0) \tag{119}$$

*Proof.* First notice that KTD is indeed a specific form of Sigma-Point Kalman Filter (SPKF). According to van der Merwe (2004, ch. 4.5), under the given assumptions, the SPKF estimator (and thus the KTD one) is the maximum a posteriori (MAP) estimator:

$$\hat{\theta}_{i|i} = \hat{\theta}_i^{\text{MAP}} = \underset{\theta}{\text{argmax}}\, p(\theta|r_{1:i}) \tag{120}$$

By applying the Bayes rule, the posterior distribution $p(\theta|r_{1:i})$ can be written as the (normalized) product of the likelihood $p(r_{1:i}|\theta)$ and of the prior distribution $p(\theta)$:

$$p(\theta|r_{1:i}) = \frac{p(r_{1:i}|\theta)p(\theta)}{p(r_{1:i})} \tag{121}$$

The normalization factor $p(r_{1:i})$ does not depend on parameters, MAP thus reduces to likelihood times prior:

$$\hat{\theta}_{i|i} = \underset{\theta}{\text{argmax}}\, p(r_{1:i}|\theta)p(\theta) \tag{122}$$

Recall that, for KTD, the observation noise is assumed white. Therefore, the joint likelihood is the product of local likelihoods:

$$\hat{\theta}_{i|i} = \underset{\theta}{\text{argmax}}\, p(r_{1:i}|\theta)p(\theta) = \underset{\theta}{\text{argmax}} \prod_{j=1}^{i} p(r_j|\theta)p(\theta) \tag{123}$$

Moreover, noise and prior are supposed to be Gaussian, thus:

$$r_j|\theta \sim \mathcal{N}\left(g_{t_j}(\theta), P_{n_j}\right) \text{ and } \theta \sim \mathcal{N}(\theta_0, P_0) \tag{124}$$

On the other hand, maximizing a product of densities is equivalent to minimizing the sum of the negatives of their logarithms:

$$\hat{\theta}_{i|i} = -\operatorname*{argmin}_{\theta} \left( \sum_{j=1}^{i} \ln(p(r_j|\theta)) + \ln(p(\theta)) \right) \qquad (125)$$

Under the Gaussian assumption, distributions are as follows:

$$p(r_j|\theta) = \frac{1}{\sqrt{2\pi P_{n_j}}} \exp\left( -\frac{1}{2} \frac{(r_j - g_{t_j}(\theta))^2}{P_{n_j}} \right) \qquad (126)$$

$$\text{and } p(\theta) = \frac{1}{(2\pi)^{\frac{p}{2}} |P_0|^{\frac{1}{2}}} \exp\left( -\frac{1}{2} (\theta - \theta_0)^T P_0^{-1} (\theta - \theta_0) \right) \qquad (127)$$

Consequently:

$$\hat{\theta}_{i|i} = \operatorname*{argmin}_{\theta} \left( \sum_{j=1}^{i} \frac{1}{P_{n_j}} (r_j - g_{t_j}(\theta))^2 + (\theta - \theta_0)^T P_0^{-1} (\theta - \theta_0) \right) \qquad (128)$$

This proves the result. □

Some remarks of importance have to be made. First, the memoryless channel assumption does not hold for stochastic MDPs. Moreover, the form of the minimized cost function (119) strengthens the parallel drawn in Section 4.1 between KTD and squared Bellman residual minimization. Second, the chosen observation noise variance $P_{n_i}$ allows weighting samples. The evolution noise variance does not appear directly in the minimized cost function, nevertheless it empirically influences convergence and tracking abilities of the algorithm. For example, it helps handling non-stationarity and avoiding local minima. The prior $P_0$ acts as a regularization terms, this can be of help to choose it. Notice that such a regularization term also appears in the recursive form of the LSTD algorithm (eg., see Kolter & Ng, 2009). Finally, it can be shown (again, see van der Merwe, 2004, ch. 4.5) that an SPKF (and thus KTD) update is indeed an online form of a modified Gauss-Newton method, which is actually a variant of natural gradient descent. In this case, the Fisher information matrix is $P_{i|i}^{-1}$, the inverse of the variance matrix of random parameters. The natural gradient approach has been shown to be quite efficient for direct policy search (Kakade, 2001) and actor-critics (Peters, Vijayakumar, & Schaal, 2005), so it lets envision good empirical results for KTD. This is experimented in Section 7. KTD is perhaps the first reinforcement learning value (and $Q$-) function approximation algorithm (in a pure critic sense) involving natural gradient.

## 5.2 Stochastic Case

Here a convergence analysis is provided for XKTD in stochastic MDPs. Again, this theorem makes some strong assumptions, without harming the minimization of the cost function (47) when they are not satisfied.

**Theorem 3.** *Assume that posterior and noise distribution are Gaussian, as well as prior distribution (of mean $\theta_0$ and variance $P_0$). Then XKTD estimator minimizes the (weighted and regularized) square error linking state values to Monte Carlo returns:*

$$C_i(\theta) = \sum_{j=1}^{i} \frac{1}{\sigma_{j-1}^2} \left( \hat{V}_\theta(s_j) - \sum_{t=j}^{i} \gamma^{t-j} r_t \right)^2 + (\theta - \theta_0)^T P_0^{-1} (\theta - \theta_0) \tag{129}$$

*Proof.* Here again the result of van der Merwe (2004, ch. 4.5) is used. The corresponding proof is made for a random walk evolution model (that is the identity evolution matrix), however it can be easily extended to a linear evolution model. It can thus be applied to state-space model (108):

$$\hat{\mathbf{x}}_{i|i} = \hat{\mathbf{x}}_i^{\text{MAP}} = \operatorname*{argmax}_{\mathbf{x}} p(\mathbf{x}|r_{1:i}) \tag{130}$$

State-space model (108) being equivalent to state-space model (46) with the MA noise (105), the same holds for the (non-extended) parameter vector:

$$\hat{\theta}_{i|i} = \operatorname*{argmax}_{\theta} p(\theta|r_{1:i}) = \operatorname*{argmin}_{\theta} (-\ln(p(\theta|r_{1:i}))) \tag{131}$$

By applying the Bayes rule, the posterior distribution $p(\theta|r_{1:i})$ is the (normalized) product of likelihood $p(r_{1:i}|\theta)$ and prior $p(\theta)$:

$$p(\theta|r_{1:i}) = \frac{p(r_{1:i}|\theta)p(\theta)}{p(r_{1:i})} \tag{132}$$

The normalization factor $p(r_{1:i})$ does not depend on parameters, MAP therefore reduces to likelihood times prior:

$$\hat{\theta}_{i|i} = \operatorname*{argmax}_{\theta} p(r_{1:i}|\theta)p(\theta) \tag{133}$$

However, as the observation noise is no longer white, it is not possible to express the joint likelihood as the product of local likelihoods. Nevertheless, the joint likelihood is still computable. For this, a few notations are introduced. Let $V_i(\theta)$, $R_i$ and $N_i$ be the following $i \times 1$ vectors:

$$V_i(\theta) = \begin{pmatrix} \hat{V}_\theta(s_1) & \hat{V}_\theta(s_2) & \dots & \hat{V}_\theta(s_i) \end{pmatrix}^T \tag{134}$$

$$R_i = \begin{pmatrix} r_1 & r_2 & \dots & r_i \end{pmatrix}^T \tag{135}$$

$$N_i = \begin{pmatrix} n_1 & n_2 & \dots & n_i \end{pmatrix}^T \tag{136}$$

Let $\mathbf{H}_i$ be the $i \times i$ bidiagonal matrix defined as:

$$\mathbf{H}_i = \begin{pmatrix} 1 & -\gamma & 0 & \dots \\ 0 & 1 & -\gamma & 0 \\ \vdots & \ddots & \ddots & -\gamma \\ 0 & \dots & 0 & 1 \end{pmatrix} \tag{137}$$

It is easy to check that its inverse is given by:

$$\mathbf{H}_i^{-1} = \begin{pmatrix} 1 & \gamma & \dots & \gamma^{i-1} \\ 0 & 1 & \gamma & \dots \\ \vdots & \ddots & \ddots & \gamma \\ 0 & \dots & 0 & 1 \end{pmatrix} \tag{138}$$

Eventually, let $\Sigma_{N_i} = E[N_i N_i^T]$ be the variance matrix of noise $N_i$, which takes into account the coloration. Given the definition of noise $n_i$ (105), its a tridiagonal matrix given by:

$$\Sigma_{N_i} = \begin{pmatrix} \sigma_0^2 + \gamma^2 \sigma_1^2 & -\gamma \sigma_1^2 & 0 & \dots \\ -\gamma \sigma_1^2 & \sigma_1 + \gamma^2 \sigma_2^2 & -\gamma \sigma_2^2 & \vdots \\ \vdots & \ddots & \ddots & -\gamma \sigma_{i-1}^2 \\ 0 & \dots & -\gamma \sigma_{i-1}^2 & \sigma_{i-1}^2 + \gamma^2 \sigma_i^2 \end{pmatrix} \tag{139}$$

As the noise is Gaussian, the likelihood is Gaussian too, and colored because of the observation noise. Its distribution is:

$$r_{1:i}|\theta \sim \mathcal{N}(R_i - \mathbf{H}_i V_i(\theta), \Sigma_{N_i}) \tag{140}$$

Maximizing MAP is equivalent to minimizing the negative of its logarithm, so given the distribution (140) the XKTD estimator satisfies:

$$\hat{\theta}_{i|i} = \underset{\theta}{\operatorname{argmin}} \left( (R_i - \mathbf{H}_i V_i(\theta))^T \Sigma_{N_i}^{-1} (R_i - \mathbf{H}_i V_i(\theta) + (\theta - \theta_0)^T P_0^{-1} (\theta - \theta_0)) \right) \tag{141}$$

The noise variance can be rewritten according to $\mathbf{H}_i$ and to a diagonal matrix containing the residual variances:

$$\Sigma_{N_i} = \mathbf{H}_i \Sigma_i \mathbf{H}_i^T \text{ with } \Sigma_i = \operatorname{diag}(\sigma_0^2, \dots, \sigma_{i-1}^2) \tag{142}$$

Using this last equation, the XKTD estimator can be rewritten as:

$$\begin{aligned} \hat{\theta}_{i|i} &= \underset{\theta}{\operatorname{argmin}} \left( (R_i - \mathbf{H}_i V_i(\theta))^T \Sigma_{N_i}^{-1} (R_i - \mathbf{H}_i V_i(\theta)) + (\theta - \theta_0)^T P_0^{-1} (\theta - \theta_0) \right) \\ &= \underset{\theta}{\operatorname{argmin}} \left( (R_i - \mathbf{H}_i V_i(\theta))^T (\mathbf{H}_i \Sigma_i \mathbf{H}_i^T)^{-1} (R_i - \mathbf{H}_i V_i(\theta)) + (\theta - \theta_0)^T P_0^{-1} (\theta - \theta_0) \right) \\ &= \underset{\theta}{\operatorname{argmin}} \left( (\mathbf{H}_i^{-1} R_i - V_i(\theta))^T \Sigma_i^{-1} (\mathbf{H}_i^{-1} R_i - V_i(\theta)) + (\theta - \theta_0)^T P_0^{-1} (\theta - \theta_0) \right) \end{aligned} \tag{143}$$

Given the inverse (138) of the $\mathbf{H}_i$ matrix, this last equation proves the result. □

This result shows that under some (strong) assumptions, XKTD minimizes the square error linking state values to Monte Carlo returns, which strengthens the discussion about the inability of XKTD to be used in an off-policy learning scenario of Section 4.3.2. As for KTD, residuals' variance weights the samples, and the prior acts as a regularization term, which can help to choose it. An important fact is that this result shows that actually, under the assumption that residuals variance is constant (that is $\sigma_j^2 = \sigma^2$), XKTD minimizes the same cost-function as (the recursive version of) LSTD(1), the eligibility traces-based extension of LSTD with and eligibility factor of 1 (see Boyan, 1999 for a proof that LSTD(1) minimizes cost-function (129)). As a consequence, XKTD is asymptotically an unbiased value function estimator, as LSTD(1)[10].

## 6. An Active Learning Scheme

The parameters being modeled as random variables, and the value (or $Q$-) function being a function of these parameters, it is a random variable for a given state (or state-action pair). It is first shown how to compute its expectation and the associated uncertainty thanks to the unscented transform. The dilemma between exploration and exploitation should benefit from such uncertainty information. Few approaches in the literature allows handling the value function approximation problem as well as computing uncertainty over values meantime. The work of Engel (2005) is such an approach, however the effective use of the obtained uncertainty information is left for future work. Here is a proposed form of active learning which is a sort of totally explorative policy in the context of KTD-Q. This contribution is shown to effectively speed up learning in Section 7.

### 6.1 Computing Uncertainty over Values

Let $\hat{V}_\theta$ be the approximated value function parameterized by the random vector $\theta$ of mean $\bar{\theta}$ and variance matrix $P_\theta$. Let $\bar{V}_\theta(s)$ and $\hat{\sigma}^2_{V_\theta}(s)$ be the associated mean and variance for a given state $s$. In order to propagate the uncertainty from the parameters to the value function, a first step is to compute the sigma-points associated to the parameter vector $\Theta = \{\theta^{(j)}, 0 \leq j \leq 2p\}$ as well as corresponding weights $\mathcal{W} = \{w_j, 0 \leq j \leq 2p\}$ from $\bar{\theta}$ and $P_\theta$, as described in Section 3.2. Then the images of these sigma-points are computed for the given state $s$ using the parameterized value function :

$$\mathcal{V}_\theta(s) = \left\{\hat{V}_\theta^{(j)}(s) = \hat{V}_{\theta^{(j)}}(s), \quad 0 \leq j \leq 2p\right\} \tag{144}$$

Knowing these images and corresponding weights, it is possible to compute the statistics of interest, namely mean and variance of the approximated value function:

$$\bar{V}_\theta(s) = \sum_{j=0}^{2p} w_j \hat{V}_\theta^{(j)}(s) \quad \text{and} \quad \hat{\sigma}^2_{V_\theta}(s) = \sum_{j=0}^{2p} w_j \left(\hat{V}_\theta^{(j)}(s) - \bar{V}_\theta(s)\right)^2 \tag{145}$$

Thus, for a given representation of the value function and a random parameter vector, the uncertainty can be propagated to the value function. Figure 1 illustrates the uncertainty computation. Extension to $Q$-function is straightforward. The complexity (both computational and in memory) is here again quadratic. So, as at each time-step $i$ an estimate $\hat{\theta}_{i|i}$ and the associated variance $P_{i|i}$ are known, uncertainty information can be computed in the KTD framework.

An important remark has to be made here. The estimated variance provides some information about the uncertainty about estimates, however it does not take into account the stochasticity of the MDP. It will get lower as the number of samples increases. Roughly speaking, it can be seen as an indirect and generalized counting of the number of visits of a given state or state-action pair. Even in a stochastic MDP, it will vanish to zero as the number of samples grows to infinity: it is an estimate of the uncertainty over the estimated value function, not the variance of the stochastic process from which the value function is the expectation.

---

10. Notice that if LSTD(1) and KTD minimize the same cost function, they do it in a different way, thus they provide the same estimates only asymptotically.

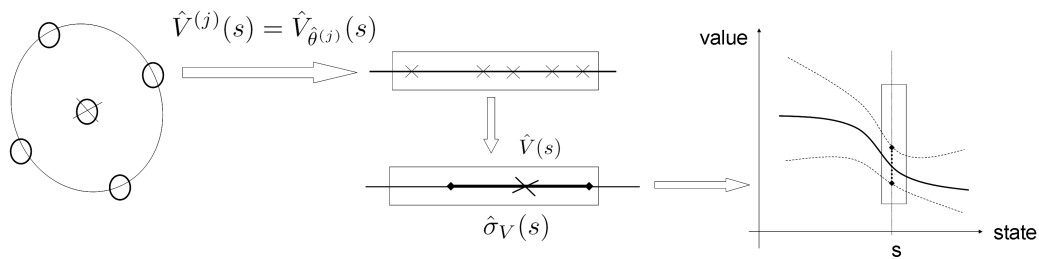

Figure 1: Uncertainty computation.

## 6.2 A Form of Active Learning

A simple active learning scheme using this uncertainty information is provided here. KTD-Q (determinism of transitions is assumed here) is an off-policy algorithm: it learns the optimal policy $\pi^*$ while following a different behavioral policy $b$. A natural question is to know what behaviorial policy to choose in order to speed up learning. A piece of response is given here.

Let $i$ be the current temporal index. The system is in a state $s_i$, and the agent has to choose an action $a_i$. The considered algorithm being KTD-Q, the estimates $\hat{\theta}_{i-1|i-1}$ and $P_{i-1|i-1}$ are available. They can be used to approximate the uncertainty of the $Q$-function parameterized by $\theta_{i-1}$ in the state $s_i$ and for any action $a$. Let $\sigma^2_{Q_{\theta_{i-1}}}(s_i, a)$ be the corresponding variance. The action $a_i$ is chosen according to the following random behaviorial policy:

$$b(a_i|s_i) = \frac{\sigma_{Q_{\theta_{i-1}}}(s_i, a_i)}{\sum_{a \in A} \sigma_{Q_{\theta_{i-1}}}(s_i, a)} \tag{146}$$

A totally explorative policy is obtained, in the sense that it favorises less certain actions. This is a way among others to use the available uncertainty information, nevertheless it is shown in Section 7 to be quite efficient compared to a uniformly random behaviorial policy. However, how to use wisely this variance information in the more general dilemma between exploration and exploitation is still an open perspective.

## 7. Experiments

This section provides a set of classical RL benchmarks aiming at comparing KTD and variants to state-of-the-art algorithms and at highlighting its different aspects. "Atomic" benchmarks have been chosen in order to highlight separately unitary properties of KTD (see Table 1), which should have been quite complex on a more difficult task. Compared algorithms are TD, SARSA and $Q$-learning with function approximation as well as (recursive form of) LSTD and (MC-) GPTD. For the sake of reproducibility, all parameter values are provided for each experiment. Their extensions to eligibility traces are not considered here, as LSTD performs better than TD($\lambda$) and varying $\lambda$ has small effect on LSTD($\lambda$) performances, according to Boyan (1999).

| | (non)stationarity | (non)linearity | uncertainty | sample efficiency | stochasticity |
|---|---|---|---|---|---|
| Tsitsiklis chain | | X | | | |
| Boyan chain | X | | | X | X |
| maze | | | X | | |
| inverted pendulum | | X | X | X | |

Table 1: Experiments and highlighted properties.

## 7.1 Choosing KTD Parameters

In order to use the (X)KTD framework, parameters have to be chosen: the variance of the observation noise (or the variance of residuals for XKTD), the priors and the variance of the process noise. As they are less common and perhaps less intuitive than the choice of a learning rate for example, they are discussed here. The evolution noise for KTD and the residual for XKTD translate the confidence the practitioner has in the ability of the chosen parameterization to represent the true value function. If it is known in advance that the value function lies in the hypothesis space (which is the case for example in the tabular case), the corresponding variance can be chosen very small (but never zero for numerical stability reasons). Another way to choose these variances is to interpret them through their weighting of samples, see Eq. (119) and (129). The prior $\theta_0$ should be initialized to a value close to the one the user thinks to be optimal, or to a default value, for example the zero vector. The prior $P_0$ quantifies the certainty the user has in the prior $\theta_0$, the lower the less certain. Another way to interpret these priors is to consider them as regularization terms, as shown in Eq. (119) and (129). How to choose the process noise variance is an open question. If some knowledge about non-stationarity is available, it can be used to choose this matrix. However, such a knowledge is generally difficult to obtain beforehand. In this article, a process noise of the form $P_{v_i} = \eta P_{\theta_{i-1|i-1}}$ is used, with $\eta \ll 1$ a small positive constant. Such an artificial process noise emphasizes recent observed data, the window of emphasized observations being quantified by $\eta$. Other artificial process noise can be chosen, see the work of van der Merwe (2004, ch. 3.5.2) for a quick survey. In the following, parameters are chosen by trial and error (for all algorithms). They're perhaps not the best ones, but orders of magnitude are correct.

## 7.2 Tsitsiklis Chain

This first experiment aims at illustrating the ability of KTD to handle nonlinear parameterizations and its convergence property. It consists in a 3 states valued Markov chain first proposed by Tsitsiklis and Roy (1997). State $i$ transits to state $i$ with probability 0.5 and to state $i-1$ with probability 0.5 too (state 1 transiting to state 1 or 3 with equi-probability). The reward is always zero, therefore the optimal value function is zero. This chain is very simple, however a nonlinear parameterization which causes TD with function approximation divergence is considered. Let $\epsilon = 0.05$, let $I$ be the $3 \times 3$ identity matrix and $M$ the

$3 \times 3$ matrix defined as:

$$M = \begin{pmatrix} 1 & \frac{1}{2} & \frac{3}{2} \\ \frac{3}{2} & 1 & \frac{1}{2} \\ \frac{1}{2} & \frac{3}{2} & 1 \end{pmatrix} \tag{147}$$

The value function is parameterized by a single scalar $\theta$, its parameterization is given as (notice that here $\hat{V}_\theta$ is a $3 \times 1$ vector):

$$\hat{V}_\theta = \exp\left((M + \epsilon I)\,\theta\right) V_0 \text{ with } V_0 = \begin{pmatrix} 10 & -7 & -3 \end{pmatrix}^T \tag{148}$$

This parameterization has been proposed by Tsitsiklis and Roy (1997) to illustrate the possible divergence of TD in the case of nonlinear parameterization. The optimal parameter is obviously $\theta^* = -\infty$.

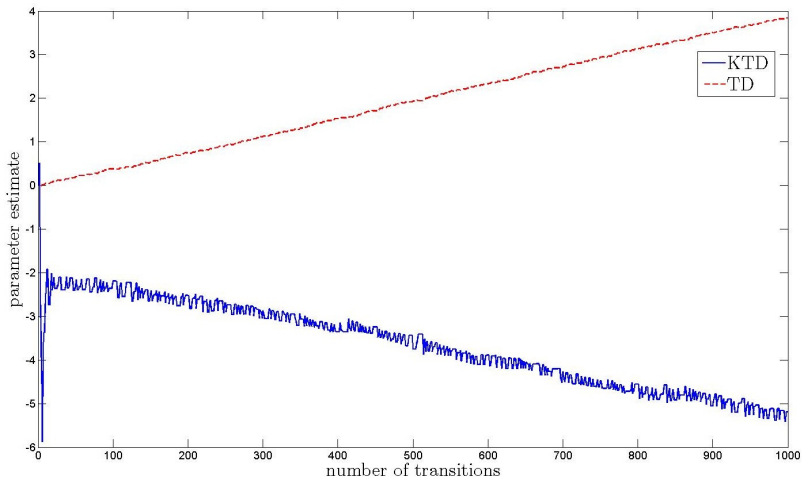

Figure 2: Tsitsiklis chain.

KTD is compared to TD with function approximation. LSTD and GPTD are not considered here, as they are unable to handle a nonlinear parameterization. For TD, the learning rate is chosen equal to $\alpha_i = 2.10^{-3}$ and the initial parameter is set to $\theta_0 = 0$. For KTD, priors are set to $\theta_0 = 0$ and $P_0 = 10$. The observation noise variance is set to $P_{n_i} = 10^{-3}$. The process noise described in Section 7.2 is used with $\eta = 10^{-1}$. Results are depicted in Figure 2 which shows the parameter estimates in function of the number of observed transitions. TD estimates diverge, as expected. KTD handles the nonlinear parameterization and converges toward the good value (despite stochasticity of transitions).

## 7.3 Boyan Chain

In this section KTD and XKTD are compared to two other second order value function approximation algorithms, namely (recursive) LSTD and (parametric) MC-GPTD on a simple valued Markov chain, the Boyan (1999) chain. The objective is threefold: showing sample efficiency, demonstrating the bias removal (of XKTD compared to KTD) and showing non-stationarity handling.

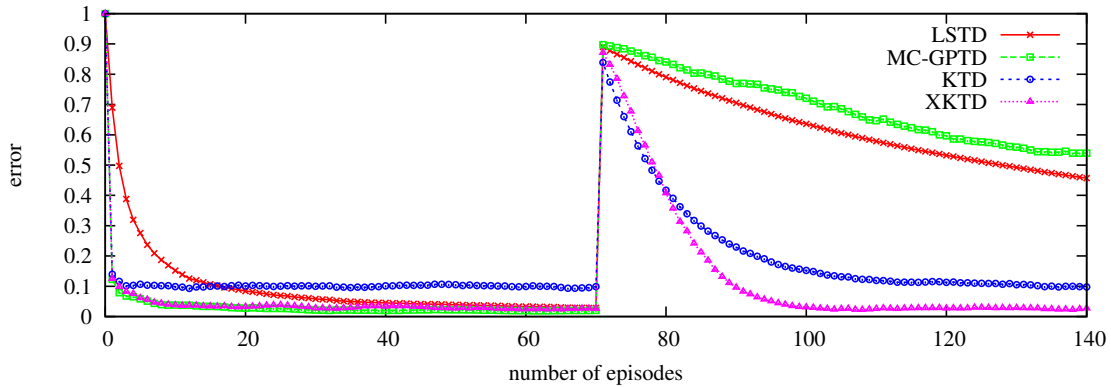

Figure 3: Boyan chain.

The Boyan chain is a 13-state Markov chain where state $s^0$ is an absorbing state, $s^1$ transits to $s^0$ with probability 1 and a reward of -2, and $s^i$ transits to either $s^{i-1}$ or $s^{i-2}$, $2 \leq i \leq 12$, each with probability 0.5 and reward -3. The feature vector $\phi(s)$ for states $s^{12}$, $s^8$, $s^4$ and $s^0$ are respectively $[1, 0, 0, 0]^T$, $[0, 1, 0, 0]^T$, $[0, 0, 1, 0]^T$ and $[0, 0, 0, 1]^T$. The feature vectors for other states are obtained by linear interpolation. The approximated value function is thus $\hat{V}_\theta(s) = \theta^T \phi(s)$. The optimal value function is exactly linear in these features, and the corresponding optimal parameter vector is $\theta_1^* = [-24, -16, -8, 0]^T$. To measure the quality of each algorithm the normalized Euclidian distance between the current parameter vector estimate and the optimal one $\frac{1}{\|\theta^*\|} \|\theta - \theta^*\|$ is computed. Notice that as the parameterization is linear, it is the same as measuring the error between the true and the estimated value functions, up to a scaling factor. The discount factor $\gamma$ is set to 1 in this episodic task. For all algorithms, the prior is set to $P_{0|0} = I$ where $I$ is the identity matrix. Choosing the same prior should be fair, as it yields to choose the same regularization term for all algorithms. For MC-GPTD and KTD variations, the residual variance (observation noise for KTD) is set to $\sigma_i^2 = 10^{-3}$ ($P_{n_i} = 10^{-3}$). For KTD variations, the process noise covariance is set to an RLS (recursive least-squares)-like adaptive process noise as described in Section 7.1, that is $P_{v_i} = \eta P_{\theta_{i-1|i-1}}$ where $P_{\theta_{i-1|i-1}}$ denotes the variance over parameters, and $\eta \ll 1$ is a small positive constant, chosen here equal to $10^{-2}$. Choosing these parameters requires some practice, but no more than choosing a learning rate for other algorithms. For all algorithms the initial parameter vector is set to zero. To experiment non-stationarity handling, a change in the MDP is simulated by multiplying the rewards by ten from the 70[th] episode (rewards become $-20$ and $-30$ instead of $-2$ and $-3$). The optimal value function is still linear in the feature vectors, and the optimal parameter vector is $\theta_2^* = 10\theta_1^*$ after the MDP change. Learning is done over 140 episodes, and results are averaged over 300 trials. Results are presented in Figure 3.

Before the MDP change, KTD variations and MC-GPTD converge faster than LSTD (and equally well). XKTD, as well as LSTD and MC-GPTD, is unbiased, contrary to KTD. Thus XKTD does the job it has been designed for, that is removing the bias due to stochastic transitions. After the MDP change, both LSTD and MC-GPTD fail to track the value function. KTD manages to do it, but it is still biased. XKTD tracks the value function without being biased. GPTD results are not presented here for the sake of readability.

However, its behavior is the same as KTD one before the MDP change, and it fails to track the value function after the rewards switch (much like MC-GPTD). This experiment shows that XKTD performs as well as KTD, however without the bias problem, which was the motivation for introducing this new algorithm. It is sample-efficient and it tracks the value function rather than converging to it (non-stationarity handling). It can be argued that some forgetting factors can be added to LSTD or GPTD. However it is more naturally done in the KTD framework, which moreover exhibits some other interesting aspects as illustrated in the next sections.

## 7.4 Simple Maze

With the KTD framework, the parameters are modelled as random variables. Being a function of the parameters, the approximated value (or $Q$-) function is a random function. It is thus possible to compute a variance associated to the value of each state as shown in Section 6.1. It is a necessary condition to handle the exploration-exploitation dilemma in a value (or $Q$-) function approximation context. In this section the uncertainty information which can be obtained from the KTD framework is illustrated on a simple maze problem.

The 2d continuous state space is the unit square: $(x, y) \in [0, 1]^2$. Actions are to move left, right, up or down, the magnitude being of 0.05 in each case. The reward is $+1$ if the agent leaves the maze in $y = 1$ and $x \in [\frac{3}{8}, \frac{5}{8}]$, $-1$ if the agent leaves the maze in $y = 1$ and $x \in [0, \frac{3}{8}[\cup]\frac{5}{8}, 1]$, and 0 elsewhere. The algorithm is KTD-V. The parameterization is a set of 9 equispaced Gaussian kernels (centered in $\{0, 0.5, 1\} \times \{0, 0.5, 1\}$) and with a standard deviation of 0.5. The forgetting factor $\gamma$ is set to 0.9. The agent starts in a random position $(x_0, y_0)$ with $x_0$ sampled from a Gaussian distribution, $x_0 \sim \mathcal{N}(\frac{1}{2}, \frac{1}{8})$, and $y_0$ sampled from a uniform distribution, $y_0 \sim \mathcal{U}_{[0, 0.05]}$. The behavorial policy for which the value function is learnt is going up with probability 0.9, and go in one of the three other directions with probability $\frac{0.1}{3}$. The initial parameter vector is set to zero, the prior to $P_{0|0} = 10I$, and the noise covariances to $P_{n_i} = 1$ and $P_{v_i} = 0I$.

The value function is learnt quite well, however this is not the point here. The objective is to illustrate the value function uncertainty. The learning is done over 30 episodes, and results are given in Figure 4, which shows the standard deviation of the approximated value function over the state space. Considering the $x$-axis, the uncertainty is lower in the middle than in the border. This is explained by the fact that learning trajectories occur more frequently in the center of the domain. Considering the $y$-axis, the uncertainty is lower near the upper bound ($y = 1$) than near the lower bound ($y = 0$). This is explained by the fact that retro-propagated values are less certain. Thus the uncertainty information computed by KTD-V is meaningful on this simple example, and it should be useful to speed up learning, eg., for exploration/exploitation dilemma. Another application example is given in the Section 6.2 and is experimented in Section 7.5. GPTD also provides a meaningful uncertainty information (Engel, Mannor, & Meir, 2003). However, as far as we know, it has never been used practically. Most likely, such uncertainty information cannot be derived from LSTD (the main reason for this belief is that the matrix maintained by LSTD is not symmetric, therefore it cannot be interpreted as a variance matrix).

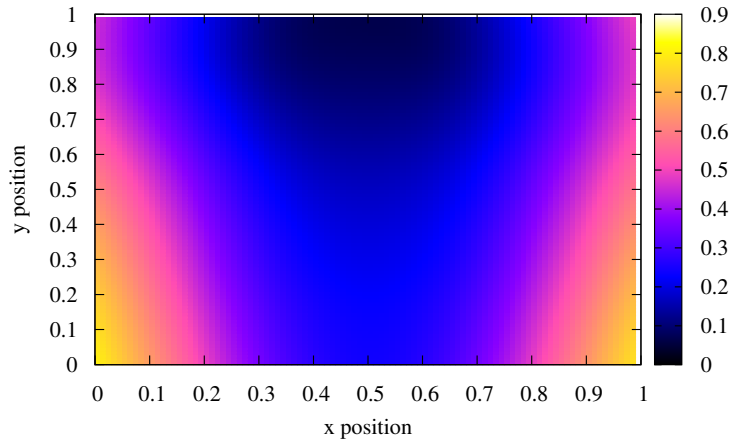

Figure 4: Simple maze, uncertainty illustration.

## 7.5 Inverted Pendulum

The last experiment is the inverted pendulum as described by Lagoudakis and Parr (2003). The goal is here to compare two value-iteration-like algorithms, namely KTD-Q and Q-learning, which aim at learning directly the optimal policy. LSTD and GPTD cannot be considered here: as they are unable to handle nonlinearities (the nonlinearity being the max operator here), they cannot be used with the Bellman optimality operator. The proposed active learning-like scheme is also experimented: it uses the uncertainty computed by KTD to speed up convergence.

This task requires balancing a pendulum of unknown length and mass at the upright position by applying forces to the cart it is attached to. Three actions are allowed: left force (-1), right force (+1), or no force (0). The associated state space consists in vertical angle $\varphi$ and angular velocity $\dot{\varphi}$ of the pendulum. Deterministic transitions are computed according to physical dynamics of the system, and depends on the current action $a$:

$$\ddot{\varphi} = \frac{g\sin(\varphi) - \beta m l\dot{\varphi}^2\sin(2\varphi)/2 - 50\beta\cos(\varphi)a}{4l/3 - \beta m l\cos^2(\varphi)} \tag{149}$$

where $g$ is the gravity constant, $m$ and $l$ the mass and the length of the pendulum, $M$ the mass of the cart, and $\beta = \frac{1}{m+M}$. A zero reward is given as long as the angular position is in $[-\frac{\pi}{2}, \frac{\pi}{2}]$. Otherwise, the episode ends and a reward of $-1$ is given. The parameterization is composed of a constant term and a set of 9 equispaced Gaussian kernels (centered in $\{-\frac{\pi}{4}, 0, \frac{\pi}{4}\} \times \{-1, 0, 1\}$ and with a standard deviation of 1) for each action. Thus there is a set of 30 basis functions. The discount factor $\gamma$ is set to 0.95.

### 7.5.1 Learning the Optimal Policy

First, algorithms ability to learn an optimal policy is compared. For Q-learning, the learning rate is set to $\alpha_i = \alpha_0 \frac{n_0+1}{n_0+i}$ with $\alpha_0 = 0.5$ and $n_0 = 200$, according to Lagoudakis and Parr (2003). For KTD-Q, the parameters are set to $P_{0|0} = 10I$, $P_{n_i} = 1$ and $P_{v_i} = 0I$. For all algorithms the initial parameter vector is set to zero. Training samples are collected online with random episodes. The agent starts in a randomly perturbed state close to the equilibrium $(0, 0)$ and then follows a policy that selects actions uniformly at random. The average length of such episodes was about 10 steps, and both algorithms learnt from the same trajectories. Results are summarized in Figure 5.

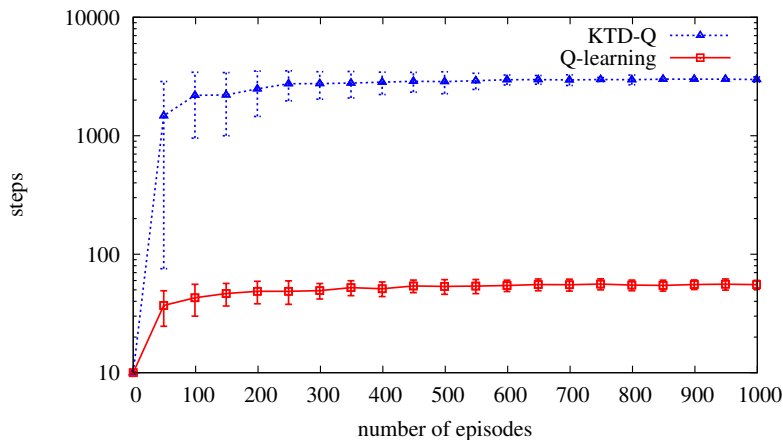

Figure 5: Inverted pendulum, optimal policy learning.

For each trial, learning is done over 1000 episodes. Every 50 episodes, learning is frozen and the current policy is evaluated. For this, the agent is randomly initialized in a state close to the equilibrium and the greedy policy is followed until the end of episode; this is repeated 100 times and averaged. Performance is measured as the number of steps in an episode. Maximum number of steps for one episode is bounded by 3000 steps, which corresponds to 5 minutes of balancing the pole without failure. Results in Figure 5 are averaged over 100 trials and presented in a semi-log scale.

KTD-Q learns an optimal policy (that is balancing the pole for the maximum number of steps) asymptotically and near-optimal policies are learnt after only a few tens of episodes. The results of KTD-Q are comparable to the ones of the LSPI algorithm (see Lagoudakis & Parr, 2003, Fig. 16). With the same number of learning episodes, $Q$-learning with the same linear parameterization fails to learn a policy which balances the pole for more than a few tens of time steps. Similar results for Q-learning are obtained by Lagoudakis and Parr (2003).

### 7.5.2 A Form of Active Learning

The parameters being random variables, as explained in Section 6 and illustrated in Section 7.4, the parameterized $Q$-function is a random function, and the KTD framework allows computing a variance associated to the value of each state. Here is proposed an experiment which aims at using this uncertainty information to speed up the learning. The learning is still done from random trajectories. However, the form of active learning described in Section 6 is considered now. The environment is initialized randomly as before. When the system is in a given state, the standard deviation of the $Q$-function is computed for each action. These deviations are normalized, and the new action is sampled randomly according to the probabilities weighted by the deviations. Thus, an uncertain action will be more likely sampled. The average length of such episodes was about 11 steps, which does not differ much from uniformly random transitions. Consequently this can only slightly help to improve speed of convergence (at most 10%, much less than the real improvement which is about 100%). Results are summarized in Figure 6.

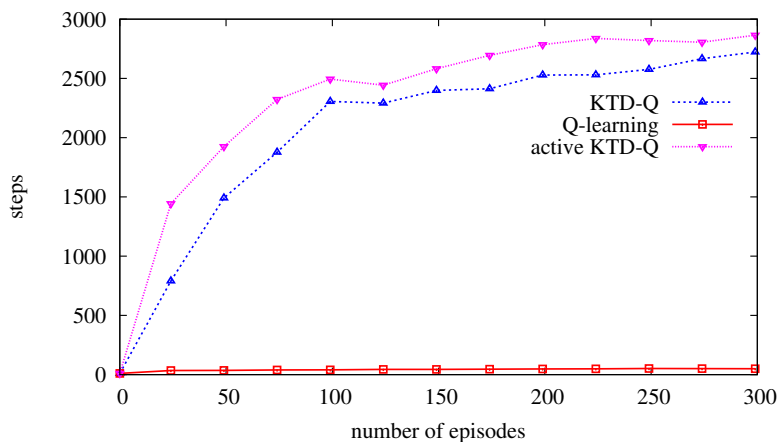

Figure 6: Inverted pendulum, random and active learning.

For each trial, learning is done over 300 episodes. Less episodes are considered to show the speed up of convergence, however both versions of KTD perform as well asymptotically. Every 25 episodes, learning is freezed and the current policy is evaluated as before. Performance is measured as the number of steps of an episode, again for a maximum of 3000 steps. Results in Figure 6 are averaged over 100 trials. Notice that the scale is no longer logarithmic. It compares KTD-Q with informed transitions ("active" KTD-Q) to KTD-Q with uniformly random learning policy and Q-learning. When comparing the two versions of KTD-Q, it is clear that sampling actions according to uncertainty speeds up convergence. It is almost doubled in the first 100 episodes: for example, a performance of 1500 is obtained after only 25 episodes with active-KTD, whereas it needs about 50 episodes for the basic KTD. Thus the uncertainty information available thanks to the KTD framework can be quite useful for reinforcement learning.

## 8. Discussion and Perspectives

In this section the proposed framework is discussed and linked to some related approaches. Some perspectives are also given.

### 8.1 Discussion

Approaches related to the KTD framework have been proposed previously. Engel (2005) proposes a Gaussian process approach to value function approximation. As explained before, its principle is to model the value function as a Gaussian process and to adopt a generative model linked to the Bellman evaluation equation. Links between Engel's approach and the proposed one have been discussed throughout the paper. Particularly, with a linear parameterization and a zero process noise KTD-V reduces to GPTD and XKTD-V to MC-GPTD. However, KTD framework handle non-stationarities (even if we recognize that GPTD could probably be extended to handle them too) and more importantly it handles non-linearities in a derivative-free manner, which allows considering nonlinear parameterizations and the Bellman optimality operator. Engel's framework allows constructing automatically and online a kernel-based linear parameterization, which is an advantage compared to the proposed framework. However, it can be easily incorporated in it (see Geist et al., 2008 where it is used in a preprocessing step, using it online is not more difficult). As Kalman filtering is strongly linked to least-squares minimization (in the linear case, the former is a generalization of the later), the proposed approach shares similarities with LSTD (Bradtke & Barto, 1996). However, it does not take into account the instrumental variables concept (Söderström & Stoica, 2002), which is used to handle stochastic transitions (in the KTD framework, it is done thanks to the colored noise model). Moreover, it has been shown in Section 5.2 that XKTD-V (with linear parameterization and no evolution noise) converges to the same solution as LSTD(1). Choi and Van Roy (2006) introduced a Kalman filter designed to handle fixed-point approximation in the case of linear parameterization. It can be roughly seen as a bootstrapping version of the proposed KTD-V. Instead of the observation equation of state-space model (65), the following observation equation is used: $r_i + \gamma \phi(s_{i+1})^T \hat{\theta}_{i-1|i-1} = \phi(s_i)^T \theta_i + n_i$. In other words, the reward is not considered as the observation, but an approximation of the value function is used to compute a "pseudo"-observation $r_i + \gamma \phi(s_{i+1})^T \hat{\theta}_{i-1|i-1}$. The update of the parameters $\theta$ is made so as to match the value function of the current state to this pseudo-observation (bootstrapping approach). Alternatively, it can be seen as a linear least-squares variation of the classic TD with function approximation algorithm (which combines bootstrapping and gradient descent). Phua and Fitch (2007) use a bank of classical Kalman filters to learn the parameters of a piecewise linear parameterization of the value function. It can be roughly seen as a special case of the proposed approach, however differences exist: not one filter but a bank is used and the parameterization is piecewise linear, which is exploited to develop specificities of the algorithm (notably concerning the parameters update) while the proposed approach does not make any assumption about the value function.

The proposed framework presents some interesting aspects. First, it does not suppose stationarity. An immediate application is to take into account non-stationary MDP (Geist, Pietquin, & Fricout, 2009b), as exemplified in Section 7.3. An even more interesting application is the control case. For instance, LSTD algorithm is known not to well behave when combined with an optimistic policy iteration scheme ($\epsilon$-greedy policy for example, see Phua & Fitch, 2007), because of the non-stationarities induced by the fact that control and learning are interlaced. Similarly, Bhatnagar, Sutton, Ghavamzadeh, and Lee (2008) prefer TD to LSTD as the actor of the incremental natural actor-critic approach they propose, despite the fact that it is less sample efficient. Kalman filtering and thus proposed approaches are robust to non-stationarity (to a certain extent). Quite few approaches aiming at approximating the value function take this non-stationary problem into account, the algorithm of Phua and Fitch (2007) being one of them. Another related approach (designed to cope with interlacing of control and learning in an actor-critic context) is the two-timescale stochastic approximation (for example, see Konda & Tsitsiklis, 2003 or Bhatnagar et al., 2008).

Second, as KTD models parameters as a random vector, it is possible to compute uncertainty information about values, as explained in Section 6.1 and illustrated in Section 7.4. It has been used to derive a form of active learning (Sections 6.2 and 7.5), however this uncertainty information could be useful to deal with the more general problem of the dilemma between exploration and exploitation, following idea of what is done by Dearden et al. (1998) or by Strehl et al. (2006). The point is that, as far as we know, rather few approaches allows dealing with value function approximation and value uncertainty in the same time. One of these approaches is the GPTD framework of Engel (2005), however the effective use of the available uncertainty information is left for future work in the original publications and has not been developed so far. It should also be noticed that without a probabilistic or statistical approach of the value function approximation problem such uncertainty information would be more difficult to obtain.

Third, KTD also allows handling nonlinearities. It has been explicitly used for KTD-Q (the max operator being a severe nonlinearity), which is illustrated in Section 7.5. Nonlinear parameterization can be considered too, as illustrated in Section 7.2. A nonlinear parameterization has also been used by Geist et al. (2008) combined with a preliminary version of KTD-Q. Moreover, nonlinear parameterization should allow more compact representation of the value function approximator, which could somehow alleviate the square complexity of the proposed framework.

KTD shares a drawback with other square Bellman residual minimization-based algorithms (which it is indeed according to Theorem 2): the value estimates are biased if transitions of the dynamic system are not deterministic, as illustrated in Section 7.3. Different algorithms propose various methods to cope with this problem. For residual algorithms (Baird, 1995), which consist in minimizing the square Bellman residual using a gradient descent, it is proposed to use double sampling in order to obtain an unbiased estimator. This approach has two major drawbacks: it needs a generative model, and it is sample inefficient. For the LSTD algorithm (Bradtke & Barto, 1996), which consists in minimizing the Bellman residual with a least-squares approach, an instrumental variable (Söderström & Stoica, 2002) is used to enforce unbiasedness of the estimator. Such an approach is not easy to extend to nonlinearity or non-stationarity (and thus online control). Another and generic approach to remove this sort of bias has been proposed by Antos et al. (2008). It consists in introducing an auxiliary function (in add to the value function) which role is to remove the bias. The resulting optimization problem is no longer quadratic, it consists in two interlocked square problems. When used with a linear function approximator, it reduces to the LSTD algorithm, and it has been used with a neural network-based function approximator by Schneegaß, Udluft, and Martinetz (2007). The GPTD framework (Engel, 2005) uses a colored noise model which has been adapted to extend the KTD framework.

## 8.2 Conclusion and Perspectives

A Kalman-filter-based Temporal Differences framework has been introduced to cope with a number of problems at the same time: online learning, sample efficiency, non-stationarity and non-linearity handling as well as providing uncertainty information. Being actually a square-Bellman-minimization-based approach, the original framework cannot handle stochastic transitions. It has thus been extended using a colored observation noise model. A convergence analysis has been provided for both deterministic and stochastic cases. Finally, various aspects of the proposed approach have been experimentally demonstrated on classical reinforcement learning benchmarks. Section 7.2 shows the ability to converge with nonlinear parameterizations, Section 7.3 shows that the colored noise induces a unbiased version of KTD and its ability to handle non-stationarities, Section 7.4 illustrates available uncertainty information and Section 7.5 shows the value-iteration-like KTD-Q algorithm as well as the learning speed-up obtained thanks to the proposed active learning scheme. State-of-the-art algorithms were also considered, and KTD compares favorably to them.

The KTD framework presents some interesting perspectives. First, XKTD was shown to effectively remove the bias. As noticed by Engel (2005, ch. 4.5), other noise models can be envisioned (by analogy to LSTD($\lambda$) for example), however what noise models to choose and how to incorporate them to the KTD framework are still open questions. More theoretical insights on the bias caused by the use of KTD on stochastic problems can also be useful. Also, an interesting perspective to address the off-policy problem when considering a colored noise is to combine XKTD with importance sampling. Another interesting perspective is to adapt the eligibility traces principle to the proposed framework in order to fill the gap between KTD (local update) and XKTD (global update by its relation to Monte Carlo) (Geist & Pietquin, 2010a).

Second, this KTD framework should be naturally extended to the partially observable case. Indeed, inferring the state of a system given past observations is a problem which can benefit from Bayesian filtering of which formalism is close to the one proposed. It is well known that a partially observable MDP (POMDP) can be expressed as an MDP of which states are distributions over states of the POMDP. If these distributions can be estimated (by using a filtering approach for example), they should be naturally taken into account by KTD: parameterization is already a function of the distribution over parameters, it can be extended to be a function of the distribution over states in the same manner.

KTD framework handles well nonlinearities. An interesting perspective could be to use it with a neural network based representation for the value (or $Q$-) function, which let hope a more compact representation. This way, it can probably be easier to address real world problems, for which scaling up is mandatory.

Another difficulty can be the choice of the different parameters, which are problem-dependent. First it should be noticed that choosing this type of parameters is not more difficult than choosing learning rates for example, it is just less usual in the RL community. Concerning a more automatic choice of parameters, the adaptive filtering literature can help (Goodwin & Sin, 2009). A form of adaptive evolution noise has been used in the experimental part of this paper, however many other solutions can be envisioned.

As said before, KTD could be an interesting alternative to TD as the actor part of the incremental natural actor-critic algorithms of Bhatnagar et al. (2008). Some preliminary works on using KTD in an actor-critic architecture are provided by Geist and Pietquin (2010c). Talking about natural gradient, a parallel has been drawn between the KTD framework and natural gradient descent in Section 5.1, and this could benefit from more theoretical insights.

The value uncertainty available from this framework has been used for a form of active learning scheme, and it is planned to be used to address the more general problem of the dilemma between exploration and exploitation, either by adapting existing approaches designed for the tabular case (Geist & Pietquin, 2010b) or by developing new methods.

Unscented Kalman filtering, on which this work is based, can be linked to nonlinear least-squares problems solved using a statistical linearization approach (Geist & Pietquin, 2010e). Underlying ideas can be used to extend the LSTD algorithm to nonlinear parameterizations as well as to the Bellman optimality operator (Geist & Pietquin, 2010d).

Finally, it is planned to do more comparison with the state-of-the-art, both theoretically and experimentally. Ultimately application of these ideas to a real world problem is needed to asses their utility. Concerning this last point, we plan to apply the proposed framework to a dialogue management problem.

## Acknowledgments

The authors wish to thank the European Community (FP7/2007-2013, grant agreement 216594, CLASSiC project : www.classic-project.org) and the Région Lorraine for financial support. Matthieu Geist also wish to thank ArcelorMittal Research for financial support during his 2006-2009 PhD thesis.